%% file: neurips_2026.tex
\documentclass{article}

\usepackage[preprint]{neurips_2026}

\usepackage[utf8]{inputenc} 
\usepackage[T1]{fontenc}    
\usepackage{hyperref}       
\usepackage{url}            
\usepackage{booktabs}       
\usepackage{amsfonts}       
\usepackage{nicefrac}       
\usepackage{microtype}      
\usepackage{xcolor}         
\usepackage{graphicx}
\usepackage{xspace}
\usepackage{booktabs}
\usepackage{tabularx}
\usepackage{array}
\usepackage{makecell}
\usepackage[normalem]{ulem}
\usepackage{enumitem}
\usepackage{amssymb}
\usepackage{mathtools}
\usepackage{caption}
\usepackage{float}
\usepackage{longtable}
\usepackage{multirow}

\usepackage[most]{tcolorbox}

\usepackage{tikz}
\usepackage[makeroom]{cancel}
\usepackage{xcolor}

\providecommand{\iflatexml}{\iffalse}

\iflatexml

\else
    
\fi

\newcommand{\methodname}{I\textit{X}T\xspace}
\newcommand{\ntp}{NTP\xspace}

\title{Introspective $X$ Training: Feedback Conditioning Improves Scaling Across all LLM Training Stages}

\renewcommand{\thefootnote}{\fnsymbol{footnote}}

\author{%
  Brandon Cui$^{1}$\thanks{Co-first authors.} \quad Ximing Lu$^{1,2}$\footnotemark[1] \quad Jaehun Jung$^{1,2}$ \quad Syeda Nahida Akter$^{1}$ \quad Hyunwoo Kim$^{1}$ \\
  \textbf{Yuxiao Qu}$^{1,3}$ \quad \textbf{David Acuna}$^{1}$ \quad \textbf{Shrimai Prabhumoye}$^{1}$ \quad \textbf{Yejin Choi}$^{1}$\thanks{Senior authors.} \quad \textbf{Prithviraj Ammanabrolu}$^{1,4}$\footnotemark[2] \\[0.4em]
  $^{1}$NVIDIA \quad $^{2}$University of Washington \quad $^{3}$Carnegie Mellon University \quad $^{4}$UC San Diego\\[0.4em]
    \texttt{\{bcui, ximingl\}@nvidia.com} 
}

\begin{document}

\maketitle

\renewcommand{\thefootnote}{\arabic{footnote}}
\setcounter{footnote}{0}

\begin{abstract}
We tackle the question of how to scale more efficiently across the many, ever-growing stages of current LLM training pipelines.
Our guiding intuition stems from the fact that the dynamics of later stages of the pipeline, e.g. post-training, can be used to inform earlier stages such as pre-training. To this end, we propose Introspective Training (or \methodname), inspired by offline reward-conditioned reinforcement learning and applicable to \textit{any} stage of training.
\methodname uses a thinking reward model to annotate data with natural language critique based feedback, enabling quality aware training from the earliest stages of the pipeline.
Models are then trained by prefix-conditioning the data with the generated feedback---ensuring that not all tokens are treated equally starting much earlier in training than usual.
Comprehensive experiments on 7.5-12B transformer-based dense LLMs trained from scratch all the way up to 18 Trillion tokens seen show that our method: \textbf{bends scaling curves resulting in up to 2.8x more compute efficiency generally}; and \textbf{reaches performance levels unachievable for models trained otherwise in domains such as math and code.} 
\end{abstract}

\input{sections/1_introduction}
\input{sections/related_work}
\input{sections/2_methods}

\input{sections/3_experimental_details}

\input{sections/4_results}
\input{sections/5_conclusion}

\newpage
\bibliographystyle{unsrtnat}
\input{Appendix/F_Limitations_Future_Work}
\bibliography{references}

\newpage
\appendix
\input{Appendix/A_annotation} 
\input{Appendix/method_details}
\input{Appendix/C_mixture}
\input{Appendix/B_MBPP}

\input{Appendix/D_IXT_scores}
\input{Appendix/E_bucketing}
\input{Appendix/G_SFT_Scores}

\end{document}

%% file: sections/1_introduction.tex
\begin{figure}[h]
    \centering
    \includegraphics[width=\textwidth]{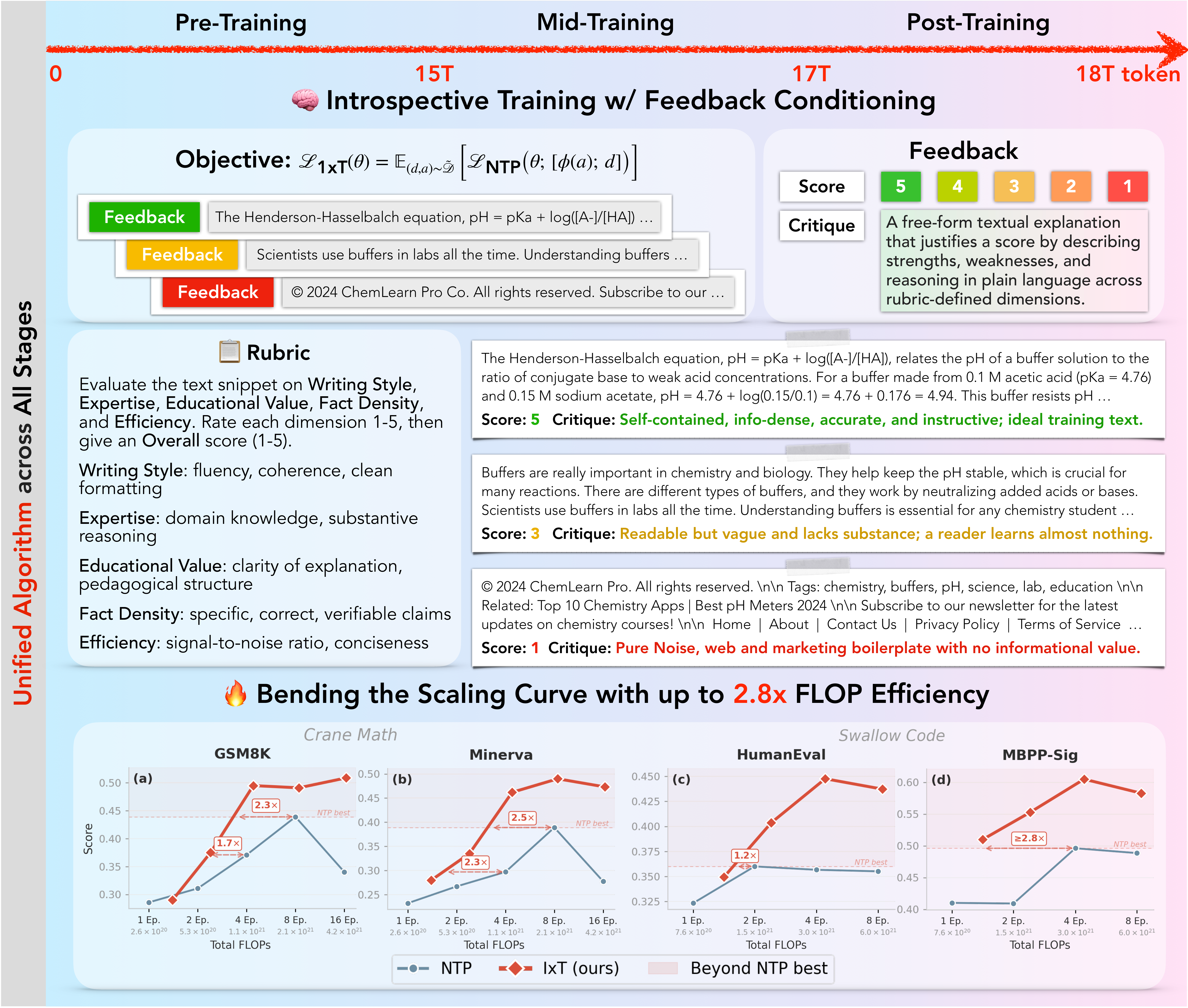}
    \caption{\textbf{Introspective Training (\methodname)}. A unified feedback-conditioning algorithm applied across pretraining, mid-training, and post-training. A judge LLM scores each document along various rubric dimensions and produces a natural language critique. Models are trained with this feedback prepended to documents, bending scaling curves with up to $2.8\times$ FLOP efficiency.}
    \label{fig:figure1}
\end{figure}

\section{Introduction}

Early LLM training pipelines were split into two stages, ``pre'' and ``post'' training.
Pre-training involved learning from massive amounts of data on the internet~\cite{hoffmann2022empirical} and post-training exposed models to data distributions more consistent with human interaction~\cite{alpaca}.
More recent variants such as ``mid'' training have sprung up to bridge the gap between passive pre-training and more active post-training, aimed at everything acquiring reasoning abilities~\cite{tan2026selfimprovingpretrainingusingposttrained} to domain specialization~\cite{blakeney2024does}. 
As training runs get longer, the number of such stages only increases, resulting in a complex multi-step process~\cite{deepseek-r1,yang2025qwen3technicalreport,blakeman2025nvidia}. 
The stages are developing semi-independently with their own datasets, algorithms, tech stacks, and evaluations---siloing knowledge and limiting cross-pollination of ideas across LLM training research. 

These standard practices are relatively unidirectional from pre-training to post-training, allowing for limited opportunities to use information from later phases to inform and enhance earlier ones.
Contemporary work also shows that what makes for a good model with respect to pre-training evaluations (such as high MMLU scores) does not necessarily help with downstream post-training evaluations\cite{hatamizadeh2025rlp,Akter2025FrontLoadingRT, Yano2026PretrainingLW}. We thus hypothesize that the pipeline could be made more computationally efficient if later-stage practices, particularly differentiating training data by quality, were introduced earlier. Rather than treating all tokens equally during pretraining only to curate more aggressively during post-training, we propose conditioning on quality signals from the start.

To this end, we propose Introspective Training, or \methodname, inspired by offline reward-conditioned reinforcement learning and illustrated in Figure~\ref{fig:figure1}.
\methodname uses existing thinking reward models~\cite{ankner2024critique,zhang2025generative} that use a rubric to annotate data with feedback---in the form of templated reward tokens or natural language critiques.
This can be done on datasets from any stage of training that uses a next-token prediction objective.
Models are then trained by conditioning existing data documents with this feedback operating effectively as a prefix, differentiating raw existing documents by not treating them all equally---especially during early (``pre'') training, something that is usually reserved for later stages. The rubric is domain agnostic by default, but the framework accommodates more targeted annotation. At test time, models are given a prefix describing desired output characteristics, either a quality label or a natural language critique. For critiques the prefixes can be refined to steer generation towards specific domains or quality dimensions, providing a pathway to use additional test-time compute.

In addition to our implementation of \methodname, we provide a comprehensive set of experiments structured around answering four research questions:

\begin{enumerate}[label=\textbf{RQ\arabic*:}]
    \item How effective is the algorithm on various stages of the LLM pipeline? 
    We provide results on transformer-based dense LLMs with at least 7 billion parameters starting from scratch, and 
    again after having seen 95B(illion), 12T(rillion), 18T and during SFT stages. 
    The results show \methodname \textbf{bends the scaling curve and shows up to 2.8x FLOP efficiency gains} 
    even accounting for additional compute used to annotate data, especially earlier during training---indicating that we can spend compute more efficiently offline once ahead of time and use fewer FLOPs during training.

    \item Can \methodname be used to domain-specialize models in a more computationally efficient manner? 
    We provide results on models that are specialized for math and code early and late in training. We show that early in training, after 95B tokens of pretraining, \methodname \textbf{delivers substantial gains ($+9.2$ on HumanEval and $+10.1$ on MATH)} \textbf{that persist at later stages in training, after 18T tokens of pretraining, up to $+10.3$ on HumanEval}.

    \item Do domain specialization results generalize to a broader set of domains 
    simultaneously? Using \methodname on a broader more general blend  \textbf{improves the overall average by $+6.3$ over standard training, with gains of up to $+12$ on math, $+13.3$ on coding, and $+2.9$ on general knowledge benchmarks,} when compared to standard training 
    

    \item Is natural language feedback effective for introspection? 
    Our north-star 
    is natural language feedback similar to human-written (and interpretable) critiques, complemented by a version that conditions on such feedback condensed into a small sequence of templated quality tokens. \textbf{Free-form critiques are $+2.6$ points better than alternative feedback forms on average}, with the templated tokens also performing better than having no conditioning.
    
\end{enumerate}

%% file: sections/related_work.tex
\vspace{-0.2cm}
\section{Related Work} \label{sec:related_work}
\vspace{-0.2cm}
\textbf{Data quality, selection, and mixture design.} Prior work improves language model training by using signals of \emph{data quality}, utility, or task relevance to shape the pretraining distribution, data mixture, or sampling schedule. This includes quality-aware document selection methods such as QuRating~\citep{wettig2024quratingselectinghighqualitydata} and Meta-rater~\citep{zhuang2025meta}, domain- and mixture-reweighting approaches such as DoReMi~\citep{xie2023doremi}, correlation- and prediction-based data selection methods such as PPLCorr~\citep{thrush2024improving} and PreSelect~\citep{shum2025predictive}, and benchmark-targeted data selection methods such as BETR~\citep{mizrahi2025betr}. Large-scale corpus efforts such as FineWeb further highlight the importance of filtering and quality-aware curation~\citep{penedo2024fineweb}, while \citet{blakeney2024does} show that late-stage upsampling of domain data can improve downstream benchmark. Our work shares the view that not all tokens should be treated equally, but instead of only filtering, ranking, or reweighting data, we attach model-generated feedback to each document and train the model to condition on it directly. This is also related to token-level filtering~\citep{rathi2026shapingcapabilitiestokenleveldata}, but differs by preserving the original data while exposing an interpretable description of document quality during training.

\textbf{Bringing post-training signals earlier into the pipeline.} Another line of work questions the strict separation between pretraining and post-training. Pretraining Language Models with Human Preferences showed that preference-derived signals can be incorporated during pretraining, with feedback-score conditioning as an effective alternative to the usual ``pretrain first, align later'' recipe~\citep{korbak2023pretraining}; however, their conditioning targeted alignment objectives (e.g., reducing toxicity) rather than improving downstream capabilities. Relatedly, \citet{maini2025safetypretraininggenerationsafe} incorporate safety objectives into pretraining through filtering, recontextualization, refusal data, and harmfulness tags. More recent work brings reasoning or RL-style supervision earlier: RLP introduces reinforcement-style objectives into pretraining~\citep{hatamizadeh2025rlp}, Front-Loading Reasoning studies when reasoning data should be introduced~\citep{Akter2025FrontLoadingRT}, \citet{zhang2025interplay} analyze the roles of pre-training, mid-training, and RL, and \citet{xing2025pretrainzero} extend RL-style optimization to pretraining.~\citet{tan2026selfimprovingpretrainingusingposttrained} further blurs this boundary by using post-trained models to rewrite pretraining data, judge continuations during training, and add reasoning traces during mid-training. Our work follows this broader goal of unifying LLM training stages, but takes a different route: rather than changing the training procedure itself, we retain the standard language-modeling objective and inject later-stage information through feedback-conditioned documents.

\textbf{Natural-language feedback and richer supervision.} Finally, our work is related to methods that learn from \emph{natural-language feedback} or richer supervisory signals. Recent work showed that language models can learn from verbal feedback as a conditioning signal without reducing critiques to scalar rewards~\citep{luo2025language}. Other approaches use language feedback for iterative refinement or policy optimization, including ILF-style training and Text2Grad, which converts critiques into fine-grained updates~\citep{chen2024learning,wang2025text2grad}. Our quality-token variant is also related to reward-token conditioning methods such as Quark, and more broadly to sequence-modeling views of reward-conditioned learning such as Decision Transformer~\citep{lu2022quarkcontrollabletextgeneration,chen2021decision}. To the best of our knowledge, no prior work combines document-level model-generated quality feedback, natural-language critiques, and standard LM training in a single interface spanning pretraining, continued pretraining, and supervised finetuning. 

%% file: sections/2_methods.tex
\section{Introspective Training}

\textbf{Preliminaries and notation.} Let $\mathcal{D} = \{d_i\}_{i=1}^N$ denote a corpus of documents, where each $d \in \mathcal{V}^*$ is a token sequence over vocabulary $\mathcal{V}$. We train an autoregressive language model $\pi_\theta$ with parameters $\theta$ under the next-token prediction (NTP) objective. Given a sequence $z = (z_1, \dots, z_T)$, the standard loss is $\mathcal{L}_{\text{NTP}}(\theta; z) = -\sum_{t=1}^{T} \log \pi_\theta(z_t \mid z_{<t})$.
 
\subsection{The \methodname Annotation Pipeline}\label{sec:ann-pipeline}
 
Given a rubric $R$ and a judge LLM $J_R$, we map each document $d \in \mathcal{D}$ to an annotation $a = J_R(d)$, where we define the annotation $a = (\textbf{s}, q, c)$, where $\mathbf{s} = (s_1,\ldots, s_5)$ is a vector of per-axis quality scores, $q \in \{1, \dots, 5\}$ is an overall score derived from $\textbf{s}$, and and $c \in \mathcal{V}^*$ is a natural language critique. The five axes are Writing Style ($s_1$), Expertise ($s_2$), Educational Value ($s_3$), Fact Density / Accuracy ($s_4$), and Efficiency ($s_5$); the critique $c$ is a structured 1--2 sentence document analysis.


All annotations are generated using Qwen3-30B-A3B~\cite{yang2025qwen3technicalreport} as $J_R$ and are computed \emph{once} prior to training, yielding an annotated corpus $\tilde{\mathcal{D}} = \{(d_i, a_i)\}_{i=1}^N$. The full annotation prompt and schema are provided in Appendix~\ref{appendix:annotation-schema}. We note that annotation is an offline, one-time cost and is accounted for in all subsequent training runs that use the annotated data.

\textbf{Pairwise Annotation.} To refine our annotation process for small post-training datasets, we also consider a pairwise annotation procedure that is particularly useful in regimes where the data distribution is narrow and pointwise scores become compressed into a small range (e.g. heavily curated SFT mixtures). Given a corpus $\mathcal{D}$, we sample pairs of documents and prompt a LLM judge to indicate which of the two would be more useful for downstream training, with document order randomized in order to mitigate position bias. Each document participates in $k=7$ such comparisons, trading off coverage against annotation compute. We then fit a Bradley-Terry model \cite{bradley-terry} over the resulting pairwise preferences using L-BFGS \cite{Liu1989LBFGS}, yielding a continuous score per document. Finally, we map these scores to the same five-level quality labels used by the pointwise annotation via manual bucketing, producing annotations $a$ that are used by \methodname without changing the training procedure. The full pairwise prompt and bucketing details are provided in Appendix~\ref{appendix:pairwise-annotation}. 
\vspace{-12pt}

\subsection{Training Conditioning Strategies} \label{sec:conditioning}
 
We explore two mechanisms to incorporate annotations during training. In both cases, we define a conditioning function $\phi$ that maps an annotation $a$ to a token prefix, prepend $\phi(a)$ to the document $d$, and train with NTP over the full concatenated sequence:
$\mathcal{L}_{\methodname}(\theta) = \mathbb{E}_{(d, a) \sim \tilde{\mathcal{D}}} \left[ \mathcal{L}_{\text{NTP}}\big(\theta;\, [\phi(a);\, d]\big) \right],$
where $[\,\cdot\,;\,\cdot\,]$ denotes sequence concatenation. We do \emph{not} mask the prefix $\phi(a)$ from the loss, allowing the model to learn the relationship between feedback and the document. For SFT, we define each document as the entire SFT conversation and prepend the conditioning signal in the system message. Since the system message is masked from the loss during SFT, the model conditions on the feedback prefix but is not trained to predict it. We instantiate $\phi$ in two ways: templated quality tokens ($\phi_{\text{tok}}$) and natural-language critiques ($\phi_{\text{crit}}$). 

\textbf{Templated Quality Tokens.} We define $\phi_{\text{tok}}(a) = \texttt{label}(s)$,
where $\texttt{label}: \{1,\dots,5\} \to \{\texttt{[low]}, \texttt{[medium-low]}, \texttt{[medium]}, \texttt{[medium-high]}, \texttt{[high]}\}$ maps the overall score to a quantized natural language label. These labels are tokenized as text sequences (not special tokens), keeping the vocabulary unchanged, following prior best practices~\cite{lu2022quarkcontrollabletextgeneration, rathi2026shapingcapabilitiestokenleveldata}. 

\textbf{Natural Language Critiques.} To provide a richer conditioning signal, we prepend the full natural language critique $\phi_{\text{crit}}(a) = c$. Critiques extend the templated approach by capturing \emph{why} a document received its quality rating, surfacing fine-grained details that a single label cannot express. Example critiques are provided in Appendix~\ref{appendix:pointwise-annotation}.

\subsection{Inference} \label{sec:inference}
 
For templated-token models, $\mathcal{P}_{\text{tok}}$ is the set of five quality labels from Section~\ref{sec:conditioning}. 
For critique-conditioned models, we construct $\mathcal{P}_{\text{crit}} = \{c^{(1)}, \dots, c^{(K)}\}$ by prompting an LLM with representative examples from the training annotation pipeline and asking it to produce critiques describing high-quality output for the target domain. This mirrors how a practitioner would use the system at deployment: describing desired output characteristics in natural language rather than selecting from a fixed set of labels.

%% file: sections/3_experimental_details.tex
\section{Experimental Setup} \label{sec:experimental_setup}

\paragraph{Notation.} We denote a model checkpoint by $\mathcal{M}$ and write $\mathcal{T}_\phi(\mathcal{M}, \mathcal{D})$ for the checkpoint obtained by continuing training from $\mathcal{M}$ on corpus $\mathcal{D}$ under conditioning strategy $\phi \in \{\phi_{\text{tok}}, \phi_{\text{crit}}, \varnothing\}$, where $\varnothing$ denotes the unconditioned \ntp baseline. $\mathcal{T}_\phi$ optimizes the \methodname objective from Section~\ref{sec:conditioning} when $\phi \neq \varnothing$, and the standard \ntp loss otherwise.

\subsection{Base Model Pretraining}
 
Our primary base checkpoint $\mathcal{M}_{\text{base}}$ is a 7.5B-parameter decoder-only transformer trained on the Nemotron Nano v2 dataset~\cite{nvidia2025nvidianemotronnano2}, comprising approximately 6.2T tokens drawn from high-quality Common Crawl, mathematics, and code. We pretrain for 95B tokens using the Nanov2 tokenizer with a maximum sequence length of 8192 tokens. Training uses a batch size of 3.5M tokens, AdamW optimizer \cite{loshchilov2019decoupledweightdecayregularization} with weight decay 0.1 and gradient clipping 1.0, and a warmup-stable-decay (WSD) learning rate schedule \cite{hu2024minicpmunveilingpotentialsmall} with a warmup phase of 2,285 steps, a decay phase of $15\%$ of total training duration, and a peak learning rate of $6.1\times10^{-4}$. Pretraining hyperparameters follow prior large-scale training recipes~\cite{deepseekai2024deepseekllmscalingopensource}. We also use this setup when pretraining models from scratch.

\subsection{Training Configurations}
 
We apply \methodname across multiple starting checkpoints, data regimes, and training stages. Unless otherwise noted, all \methodname configurations use templated quality tokens as the conditioning signal ($\phi = \phi_{\text{tok}}$).

\textbf{Baseline: Next Token Prediction (NTP).} We train baselines on unannotated documents with the standard \ntp loss and evaluate with matched prefix selection: $\mathcal{P}_{\text{tok}} \cup \{\varnothing\}$ or $\mathcal{P}_{\text{crit}} \cup \{\varnothing\}$, depending on the \methodname variant being compared, where $\varnothing$ denotes no prefix. This isolates training-time gains from inference-time prompting benefits.
 
\textbf{Domain Specific Continued pretraining (CPT).}
We evaluate CPT in two settings. For scaling behavior, we produce $\mathcal{T}_\phi(\mathcal{M}_{\text{base}}, \mathcal{D})$ for up to 16 epochs on $\mathcal{D} \in \{\mathcal{D}_{\text{CraneMath}}, \mathcal{D}_{\text{SwallowCode}}\}$. For robustness in stronger models, we continue from 12T and 18T token checkpoints of a 12B-parameter Mamba-Transformer hybrid model (trained on the same datasource~\cite{nvidia2025nvidianemotronnano2}) for up to 4 epochs (up to ${\sim}$34B additional tokens), selecting the best checkpoint per epoch.

\textbf{General-purpose blends.}
To test generalization beyond domain-specific corpora, we continue training from $\mathcal{M}_{\text{base}}$ on Dolmino~\cite{olmo2025olmo3} for 1 epoch (\textasciitilde 106B tokens). We reweight the Dolmino mixture to upweight structured domains where quality signal is strongest (Full mixture details are provided in Appendix~\ref{app:dolmino-mix}). 
We partition Dolmino as $\mathcal{D}_{\text{Dolmino}} = \mathcal{D}_{\text{Crane}} \cup \mathcal{D}_{\text{rest}}$, where $\mathcal{D}_{\text{Crane}} = \mathcal{D}_{\text{CraneMath}} \cup \mathcal{D}_{\text{CraneCode}}$. We compare three strategies: \textit{NTP}, \textit{All Annotations} (applying $\phi$ to all of $\mathcal{D}_{\text{Dolmino}}$), and \textit{Specific Subset Annotation} (only applying $\phi$ to $\mathcal{D}_{\text{Crane}}$).



\textbf{Supervised finetuning (SFT).}
Starting from the Dolmino-trained checkpoint $\mathcal{M}_{\text{Dolmino}} = \mathcal{T}_\phi(\mathcal{M}_{\text{base}}, \mathcal{D}_{\text{Dolmino}})$, we apply SFT with the same $\phi$ used in CPT, yielding $\mathcal{T}_\phi(\mathcal{M}_{\text{Dolmino}}, \mathcal{D}_{\text{SFT}})$. We evaluate on three SFT datasets: \textit{General SFT}, \textit{OpenThinker}, and \textit{SFT $>$ 4k}.

Full training configuration details are provided in Appendix~\ref{appendix:training-details}.


\subsection{Evaluation} \label{sec:evaluation}
 
We evaluate in a zero-shot setting across benchmarks spanning mathematical reasoning (GSM8K~\cite{cobbe2021trainingverifierssolvemath}, MATH~\cite{hendrycks2021measuringmathematicalproblemsolving}), code generation (HumanEval~\cite{chen2021evaluatinglargelanguagemodels}, MBPP~\cite{austin2021programsynthesislargelanguage}), and general language understanding (MMLU~\cite{hendrycks2021measuringmassivemultitasklanguage}, MMLU-Pro~\cite{wang2024mmluprorobustchallengingmultitask}, ARC-Challenge~\cite{clark2018thinksolvedquestionanswering}, and RACE\cite{LaiRACE2017} ). For SFT models, we evaluate on AIME24/25~\cite{AIME2024, AIME2025}, AMC~\cite{AMC}, IFEval~\cite{zhou2023instruction}, OlympiadBench~\cite{he2024olympiadbenchchallengingbenchmarkpromoting}, LiveCodeBench~\cite{jain2024livecodebenchholisticcontaminationfree}, Minerva~\cite{lewkowycz2022solvingquantitativereasoningproblems}, HumanEval~\cite{chen2021evaluatinglargelanguagemodels}, MMLU~\cite{hendrycks2021measuringmassivemultitasklanguage}, and MMLU-Pro~\cite{wang2024mmluprorobustchallengingmultitask}.\footnote{Full evaluation details are in Appendix~\ref{appendix:eval-details}.} 
 
\textbf{Compute accounting.} All FLOPs comparisons account for both training compute and the one-time annotation inference cost. We follow the conventional FLOPs approximation for transformer models with $N$ parameters: $6N$ per training token and $2N$ per inference token~\cite{Kaplan2020Scaling}. For a training run processing $T_{\text{train}}$ tokens with a model of size $N_{\text{train}}$, annotated once by a judge of size $N_{\text{ann}}$ that processes $T_{\text{ann}}$ total tokens, comprising of both input documents (prefill) and generated critiques and scores, the total FLOPs are:

\begin{equation}
\text{FLOPs}_{\text{total}} = \underbrace{6 \, N_{\text{train}} \, T_{\text{train}}}_{\text{training}} + \underbrace{2 \, N_{\text{ann}} \, T_{\text{ann}}}_{\text{annotation}},
\end{equation}
with the second term set to zero for the \ntp baseline. This ensures \methodname's efficiency gains are measured fairly against \ntp.

%% file: sections/4_results.tex
\section{Results} \label{sec:results}

\textbf{RQ 1: How effective is \methodname across stages of the LLM pipeline?} \label{RQ1}

\textbf{Pretraining from scratch.} We pretrain a 7.5B transformer from scratch on the reweighted Dolmino mixture for one epoch, applying \textit{specific subset annotation} such that only documents in $\mathcal{D}_{\text{crane}}$ receive quality-conditioned prefixes. As shown in table~\ref{tab:pretrain-scratch}, \methodname improves the overall average from 43.1 to 48.7 (+5.6), with the largest gains on math and code (GSM8k: $35.8 \rightarrow 50.2$, MATH: $42.2 \rightarrow 53.9$, HumanEval: $32.8 \rightarrow 37.7$, MBPP: $50.2 \rightarrow 58.4$). The improvements also extend to general purpose benchmarks (MMLU: $40.4 \rightarrow 42.7$, MMLU-Pro $19.6 \rightarrow 24.4$). This result demonstrates that the benefits of feedback conditioning can be achieved from random initialization.

\textbf{Early checkpoint performance.}
Figure~\ref{fig:figure1} plots domain-specific benchmark performance against total FLOPs (training + annotation inference) when continuing training from $\mathcal{M}_{base}$. Across all four benchmarks, GSM8k, MATH, HumanEval, and MBPP, \methodname achieves higher accuracy at matched compute, with up to 2.8$\times$ FLOPs savings at matched performance levels. \methodname without conditioning tracks the \ntp baseline closely; this confirms the conditioning signal is driving the improvement. These early-stage gains are not limited to domain-specific corpora: even on a general-purpose blend (Dolmino, Table~\ref{tab:dolmino-mix}), \methodname improves the overall average score from 46.8 to 53.1.

\textbf{Efficacy at scale.}
To test if quality conditioning applies at later model training stages, we apply \methodname at 12T and 18T token checkpoints and present results in Table~\ref{tab:nano-v2-domain-specific}. \methodname improves over \ntp across math and code benchmarks after up to 2 epochs of CPT. Importantly, \methodname applied at 12T surpasses the 18T general-corpus baseline (without domain CPT) on HumanEval (65.8\% vs.\ 59.0\%), suggesting that targeted quality-conditioned specialization at an earlier checkpoint can be more effective than simply training longer on a general corpus.

\textbf{Effectiveness during SFT.} Table~\ref{tab:SFT} extends the evaluation to SFT across three datasets: General SFT, OpenThinker, and SFT > 4k. \methodname improves the overall average on all three ($36.5\rightarrow 37.7$, $49.2\rightarrow 50.0$, and $43.7 \rightarrow 45.4$), though gains are smaller than in CPT and distributed differently across categories. We attribute the reduced magnitude to SFT operating on a narrower, more curated distribution where quality differentiation is less pronounced. On OpenThinker, where data is already very heavily filtered, the gains are concentrated on math (+1.5) with a small regression on coding (-0.9), consistent with the hypothesis that pre-curated data leaves less headroom for quality conditioning. The direction is consistently positive on overall averages, supporting the broader idea that conditioning remains useful when applied to high-quality SFT mixtures.

We also apply our pairwise annotation procedure to \textit{SFT > 4k}, see section \ref{sec:ann-pipeline}, enabling us to reach 46.1 on average, which is 2.4 points above \ntp and 0.7 point above the original annotation. This suggests that targeted compute spent calibrating annotations for SFT is a promising direction.


This cross-stage consistency, training from scratch through 18T tokens and into SFT is a key property for a unified training method. 



\begin{table}[t]
\centering
\footnotesize
\setlength{\tabcolsep}{3.5pt}
\renewcommand{\arraystretch}{1.05}

\begin{tabularx}{\linewidth}{
>{\raggedright\arraybackslash}p{3.2cm}
*{9}{>{\centering\arraybackslash}X}
}
\toprule
Method & GSM8K & MATH & \makecell{Human\\Eval} & \makecell{MBPP} & MMLU & MMLU-Pro & ARC-C & RACE & Avg. \\
\midrule
\ntp & 35.8 & 42.2 & 32.8 & 50.2 & 40.4 & 19.6 & \textbf{58.1} & 65.6 & 43.1 \\
\methodname & \textbf{50.2} & \textbf{53.9} & \textbf{37.7} & \textbf{58.4} & \textbf{42.7} & \textbf{24.4} & 56.1 & \textbf{65.8} & \textbf{48.7}\\
\bottomrule\\
\end{tabularx}
\caption{Performance of pretraining our 7.5B model from scratch with Dolmino across our evaluation suite comparing \ntp to \methodname applied to targeted subsets, \textit{Specific Subset Annotation}. \methodname outperforms \ntp across almost all benchmarks with the largest gains in more reasoning heavy benchmarks.}
\label{tab:pretrain-scratch}
\end{table}

\begin{table}[t]
\centering
\footnotesize
\begin{tabularx}{\linewidth}{
>{\raggedright\arraybackslash}p{3.2cm}
*{4}{>{\centering\arraybackslash}X}
}
\toprule
& \multicolumn{2}{c}{Math} & \multicolumn{2}{c}{Code} \\
\cmidrule(lr){2-3} \cmidrule(lr){4-5}
Tokens Seen + Method & GSM8k & MATH & HumanEval & MBPP \\
\midrule
\textit{Starting Checkpoints}\\
12T & 82.9 & 56.7 & 43.8 & 49.1 \\
18T & 92.1 & 77.6 & 59.0 & 59.6 \\
\midrule
\midrule
& \multicolumn{2}{c}{\textit{Crane Math}} & \multicolumn{2}{c}{\textit{Swallow Code}} \\
12T + \ntp & 77.9 & 60.0 & 55.2 & 62.0 \\
12T + \methodname & \textbf{84.2} & \textbf{66.9} & \textbf{65.8} & \textbf{64.5} \\
\midrule
18T + \ntp & 78.3 & 66.1 & 59.7 & \textbf{60.0} \\
18T + \methodname & \textbf{87.5} & \textbf{69.6} &  \textbf{70.0} & 55.3\\
\bottomrule\\
\end{tabularx}
\caption{Performance across math (GSM8K and MATH) and code (HumanEval and MBPP) benchmarks with continued pretraining. Models are initialized from checkpoints trained on 12T and 18T tokens, and further trained on domain-specific datasets (Crane Math for math and Swallow Code for code), with evaluation on corresponding domain benchmarks. At comparable starting checkpoints and additional training budgets, \methodname in almost all cases improves over \ntp.}
\label{tab:nano-v2-domain-specific}
\vspace{-24pt}
\end{table}

\textbf{RQ 2: Can \methodname be used to domain-specialize models more computationally efficiently?} \label{RQ2}

\textbf{\methodname Scaling behavior.} Figure~\ref{fig:figure1} plots domain-specific performance against total FLOPs, when continuing from $\mathcal{M}_{base}$ and using templated quality tokens as the conditioning signal. On several benchmarks \methodname surpasses the \ntp asymptote entirely. For instance, on HumanEval, \ntp plateaus below the accuracy \methodname achieves at a fraction of the compute. This is the sense in which \methodname breaks through ceilings that standard training cannot: on a fixed corpus, \ntp has an asymptotic performance limit that quality-conditioned training surpasses. Tables~\ref{tab:IxT-critiques} and \ref{tab:dolmino-critiques} show that these domain specialization gains are not tied to a particular feedback format (see RQ4 for a detailed comparison).

\textbf{Persistence at scale.} Table~\ref{tab:nano-v2-domain-specific} shows these improvements are not limited to early-stage models. We continue training a Mamba-Transformer hybrid model from 12T and 18T checkpoints with $\mathcal{D}_{crane}$. \methodname exceeds the performance of \ntp at both starting checkpoints. Specifically, at 12T \methodname achieves improvements of up to 10.6 points on HumanEval and 6.9 points on MATH over standard training, and at 18T \methodname improves in coding, where we see a 10.3 point increase on HumanEval but we see a 4.7 point decrease on MBPP compared to \ntp. Overall, this indicates that \methodname can still be applied at later training stages.

\begin{table}[t]
\centering
\footnotesize
\setlength{\tabcolsep}{3.5pt}
\renewcommand{\arraystretch}{1.05}

\begin{tabularx}{\linewidth}{
>{\raggedright\arraybackslash}p{3,2cm}
*{9}{>{\centering\arraybackslash}X}
}
\toprule
Method & GSM8K & MATH & \makecell{Human\\Eval} & \makecell{MBPP} & MMLU & MMLU-Pro & ARC-C & RACE & Avg. \\
\midrule \\
\ntp & 29.7 & 39.6 & 37.2 & 52.5 & 48.9 & 24.9 & 65.9 & \textbf{75.9} & 46.8 \\
\methodname w/ \textit{All Ann.} & \textbf{68.5} & 39.8 & 39.5 & 56.4 & 49.3 & \textbf{28.4} & \textbf{69.3} & 74.7 & \textbf{53.2} \\
\methodname w/ \textit{Specific Subset Ann.} & 41.2 & \textbf{51.6} & \textbf{50.5} & \textbf{60.0} & \textbf{50.8} & 27.4 & 67.8 & 75.1 & 53.1 \\
\bottomrule \\
\end{tabularx}
\caption{Performance when training $\mathcal{M}_{base}$ on Dolmino under different training strategies: training with the standard next token prediction (\textit{\ntp}), applying \methodname to the full Dolmino dataset (\textit{All Ann.}), and applying \methodname to targeted subsets (\textit{Specific Subset Ann.}). By applying \methodname to specific subsets, we are able to achieve comparable average performance to full-blend annotation while delivering the largest improvements on more reasoning intensive benchmarks.}
\label{tab:dolmino-mix}
\end{table}

\begin{figure*}[t!]
\includegraphics[width=\textwidth]{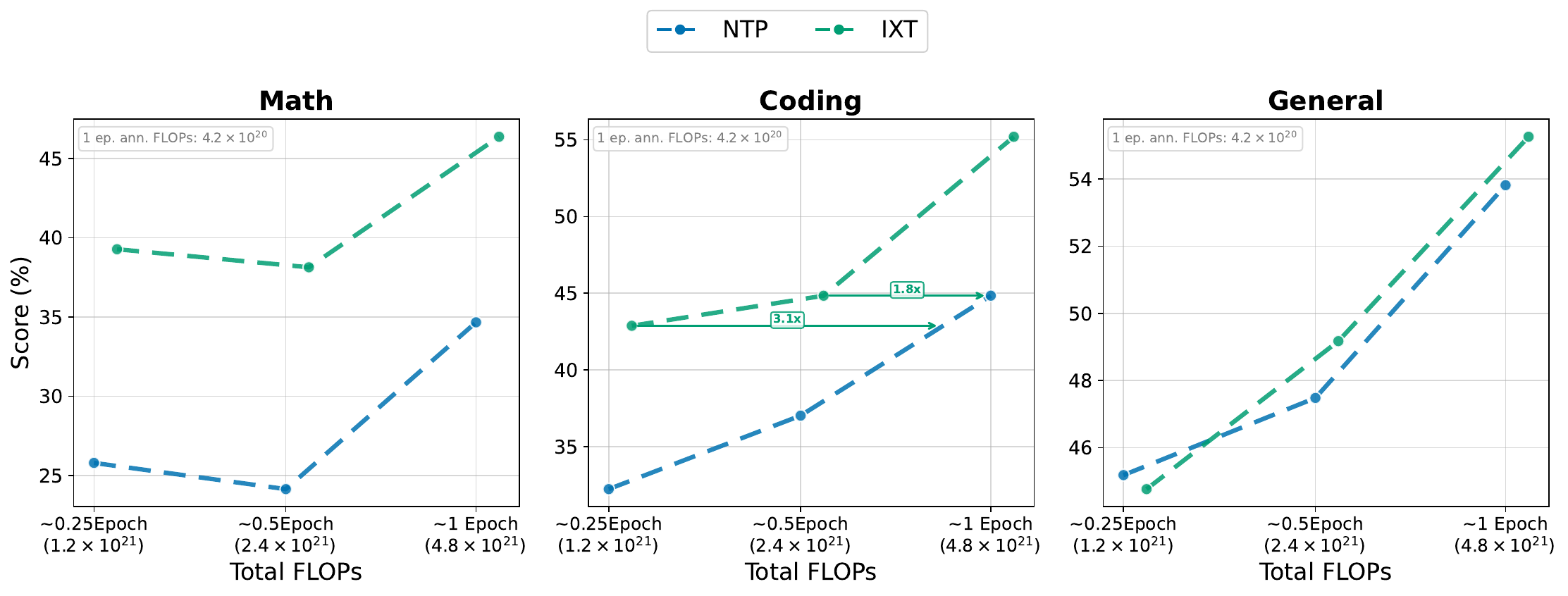}
\caption{Flop-scaling curves for \methodname vs NTP on Dolmino averaged across Math (MATH and GSM8k), Coding (HumanEval and MBPP), and General (MMLU, MMLU-Pro, Arc-Challenge, and RACE), where we apply \methodname to only \textit{Specific Subsets} of Dolmino. \methodname consistently outperforms standard next token prediction at matched compute budgets. For math, NTP results never achieve parity with \methodname, and for coding it requires 1.8-3.1x the amount of compute to do so.}
\label{fig:dolmino_scaling}
\end{figure*}

\textbf{RQ 3: Do domain specialization results generalize to a broader set of domains simultaneously?} \label{RQ3} A next natural question is whether \methodname must be applied exhaustively across a full data blend, or whether targeted annotation suffices. We test this using the Dolmino dataset, comparing \ntp, \methodname applied to all documents, and \methodname applied only to $\mathcal{D}_{crane}$.

As shown in Table~\ref{tab:dolmino-mix}, applying \methodname to targeted subsets (\textit{Specific Subset Annotation}) typically matches or exceeds the full blend annotation, while annotating only $15\%$ of the data. Full-blend annotation retains an edge on GSM8k, likely because the GSM8k train set is in Dolmino, which would directly reinforce the conditioning signal at evaluation time; on other sources such as Common Crawl, the quality signal is noisier and less well-calibrated, diluting the benefit. Targeted annotation sidesteps this by restricting \methodname to $\mathcal{D}_{crane}$, where quality differentiation is most meaningful. Beyond the annotated, domains we also observe positive spillover to general purpose benchmarks (e.g. MMLU +2.1, ARC-Challenge +1.9), suggesting that \methodname is most effective when applied selectively to domains with structured reasoning or clear quality gradients.


Figure~\ref{fig:dolmino_scaling} reinforces these findings when viewed through a computational efficiency lens. Across all three benchmark categories, \methodname with Specific Subset Annotation consistently outperforms \ntp at matched FLOP budgets. The gains are most pronounced in coding, where \methodname achieves $3.1\times$ compute efficiency over \ntp, and in math, where \methodname leads \ntp by 13.5 points at 0.25 epochs and maintains a 10.7 point advantage at 1 epoch. On general purpose, \ntp and \methodname have similar performance at $0.25$ epochs, but \methodname's advantage widens with compute, reaching a final gap of $+1.5$ points. The annotation cost itself is a small fraction of training FLOPs, confirming the overhead of targeted annotation is amortized via downstream gains.

\begin{table}[t]
\centering
\footnotesize
\begin{tabularx}{\linewidth}{
>{\raggedright\arraybackslash}p{3.2cm}
*{9}{>{\centering\arraybackslash}X}
}
\toprule
Method & Math Avg. & Coding Avg. & General Avg. & Overall Avg. \\
\midrule
\textit{General SFT} \\
\ntp & 34.1 & 36.3 & 42.0 & 36.5 \\
\methodname & \textbf{34.9} & \textbf{37.2} & \textbf{44.6} & \textbf{37.7} \\
\midrule
\midrule
\textit{Openthinker}\\
\ntp & 54.0 & \textbf{44.0} & 41.6 & 49.2 \\
\methodname & \textbf{55.5} & 43.1 & \textbf{41.8} & \textbf{50.0}\\
\midrule
\midrule
\textit{SFT > 4k} \\
\ntp & 50.6 & 37.2 & 31.8 & 43.7\\
\methodname & \textbf{52.9} & 37.6 & 33.4 & 45.4 \\
\methodname pairwise ann. & \textbf{52.9} & \textbf{38.8} & \textbf{34.7} & \textbf{46.1}\\
\bottomrule\\
\end{tabularx}
\caption{Performance after SFT, averaged for math benchmarks (AIME24, AIME25, AMC, GSM8k, Minerva MATH, OlympiadBench), Coding Benchmarks (LiveCodeBench, HumanEval), General Benchmarks (IFEval, MMLU, MMLU-Pro), and averaged over all benchmarks. We consider three different datasets \textit{General SFT}, \textit{OpenThinker}, and \textit{SFT > 4k}. For the \textit{SFT > 4k} dataset we also consider our pairwise annotation process (pairwise ann.). \methodname leads to improvements across all datasets and most categories. By spending more compute at annotation time with pairwise annotation, we are able to further improve results.}
\label{tab:SFT}
\vspace{-18pt}
\end{table}

\begin{table}[t]
\centering
\footnotesize
\begin{tabularx}{\linewidth}{
>{\raggedright\arraybackslash}p{3.2cm}
*{4}{>{\centering\arraybackslash}X}
}
\toprule
& \multicolumn{2}{c}{Math} & \multicolumn{2}{c}{Code} \\
\cmidrule(lr){2-3} \cmidrule(lr){4-5}
Method & GSM8K & MATH & HumanEval & MBPP \\
\midrule
& \multicolumn{2}{c}{\textit{Crane Math}} 
& \multicolumn{2}{c}{\textit{Swallow Code}} \\
\ntp & 43.9 & 38.9 & 35.7 & \textbf{49.6} \\
\methodname & \textbf{46.8} & \textbf{46.1} & \textbf{45.9} & 47.5 \\
\bottomrule \\
\end{tabularx}
\caption{Natural language critiques improve domain-specific performance. We compare \ntp and \methodname when using natural critiques for training and evaluation. Models are trained on (a) Crane Math and evaluated on GSM8K and MATH, and (b) Swallow Code and evaluated on HumanEval and MBPP. \methodname with critiques typically improves over \ntp on most benchmarks.}
\label{tab:IxT-critiques}
\end{table}

\begin{table}[t]
\centering
\footnotesize
\setlength{\tabcolsep}{3.5pt}
\renewcommand{\arraystretch}{1.05}

\begin{tabularx}{\linewidth}{
>{\raggedright\arraybackslash}p{3,2cm}
*{9}{>{\centering\arraybackslash}X}
}
\toprule
Method & GSM8K & MATH & \makecell{Human\\Eval} & \makecell{MBPP} & MMLU & MMLU-Pro & ARC-C & RACE & Avg. \\
\midrule \\
\ntp w/ crit. & 29.7 & 48.7 & 46.0 & 49.8 & 49.8 & \textbf{28.4} & \textbf{68.4} & 75.9 & 49.6\\
\methodname $\phi_{tok}$ & 41.2 & \textbf{51.6} & 50.5 & 60.0 & 50.8 & 27.4 & 67.8 & 75.1 & 53.1 \\
\methodname $\phi_{crit}$ &  \textbf{59.5} & 46.5 & \textbf{54.9} & \textbf{60.9} & \textbf{50.9} & 27.9 & 67.9 & \textbf{77.1} & \textbf{55.7} \\
\bottomrule \\
\end{tabularx}
\caption{Performance when training $\mathcal{M}_{base}$ on Dolmino on unannotated documents (\textit{NTP}), with \methodname with templated tokens (\methodname $\phi_{tok}$), and with \methodname with natural language critiques (\methodname $\phi_{crit}$). Natural Language critiques help on average improve further over templated tokens by $+2.6$.}
\label{tab:dolmino-critiques}
\vspace{-12pt}
\end{table}


\textbf{RQ 4: Is natural language feedback effective for introspection?} \label{RQ4}
Here we examine whether conditioning can be done effectively via natural language critiques. Critiques offer two practical advantages: 1) users can steer model behavior through descriptive language during deployment and 2) they enable naturally richer feedback as annotation pipelines improve. So, establishing their effectiveness is practically important.

Table~\ref{tab:IxT-critiques} compares \ntp and \methodname with critiques at both training and evaluation time. On Crane Math critique-conditioned \methodname improves over \ntp on both benchmarks, reaching 46.8\% on GSM8k (vs. 43.9\% for \ntp) and 46.1\% on MATH (vs. 38.9\%). On Swallow Code, critique-conditioned \methodname improves substantially on HumanEval (45.9\% vs. 35.7\% for \ntp) but underperforms \ntp on MBPP (47.5\% vs 49.6\%). We attribute the MBPP result to a rubric domain mismatch: our annotation rubric is calibrated for general-purpose text quality (see Figure~\ref{fig:pretraining-prompt}) rather than code-specific notions such as correctness, modularity, and algorithmic efficiency, which limits the critique signal in this domain. 

These gains extend beyond domain-specific CPT. On the Dolmino general-purpose blend with Specific Subset Annotation (Table~\ref{tab:dolmino-critiques}), replacing templated tokens with critiques raises the overall average from 53.1 to 55.7, with substantial gains on GSM8k (+18.3) and HumanEval (+4.4), though results are mixed on MATH (-5.1). This suggests critique conditioning remains effective in general regimes.

Natural language critiques achieve performance comparable to templated quality tokens while offering a strictly more expressive interface. Together, the two conditioning strategies form a coarse-to-fine spectrum: templated tokens provide a lightweight, robust signal, while critiques offer richer expressiveness and inference-time controllability. We hypothesize that critiques represent a more scalable form of feedback that will benefit from improvements in annotation quality, particularly as domain-adapted rubrics improve the signal for code and other domains where generic critiques are less well-calibrated.

%% file: sections/5_conclusion.tex
\vspace{-2pt}
\section{Conclusion}
\vspace{-6pt}
We introduce \textbf{Introspective Training (\methodname)}, a unified algorithm that conditions language model training on rubric-derived feedback, either templated quality tokens or natural language critiques. We apply the same recipe across different training regimes: pretraining, continued pretraining, and SFT. Across the four research questions, \methodname bends scaling curves with up to a $2.8\times$ FLOP efficiency gain, surpasses \ntp asymptotes on fixed corpora, persists when applied to 12T and 18T token checkpoints, and yields a positive spillover to general benchmarks even when only ~15\% of the data is annotated. Critique based conditioning is also competitive with templated tokens while offering a strictly more expressive interface for inference-time control. Together, these results show that letting feedback flow backwards through the training pipeline is a practical and broadly effective path toward more compute-efficient LLM training.

%% file: Appendix/F_Limitations_Future_Work.tex
\newpage
\section{Limitations and Future Work} \label{Appendix:Limitations}
While \methodname demonstrates consistent gains across training stages, our annotation rubric was designed for general-purpose text quality and may not optimally capture domain-specific notions of quality. A natural extension is to improve the rubric optimization algorithm, especially in domains where generic criteria are poorly calibrated. Additionally, we compare two conditioning formats, templated quality tokens and natural language critiques, but we do not ablate more fine-grained training-time conditioning choices such as the granularity or verbosity of the critique, different formats of the natural language critique, or mixing conditioning signals---deeming it beyond the scope of this work. 

\methodname annotations are also computed once before training and remain fixed throughout. As the model improves, quality distinctions that were informative early in training may become less discriminative. An online or iterative variant that periodically re-annotates data against the evolving model capability is another possible direction.
Overall, we view \methodname as a first step towards unifying various LLM training stages in a computationally efficient manner.

\section{Broader Impacts} \label{Appendix:Broader-Impacts}
This work proposes a method for improving the compute efficiency of LLM training by conditioning on model-generated quality feedback. More efficient training pipelines reduce the resources required to achieve a given capability level, potentially broadening access to capable language models, though this also lowers the barrier to entry to train such models with limited oversight. As a secondary benefit the annotation pipeline produces interpretable quality signals that could support data auditing and curation practices. Overall, we encourage practitioners to apply these methods following common best practices and guidelines. 

%% file: Appendix/A_annotation.tex
\section{Annotation Schema} \label{appendix:annotation-schema}
\subsection{Pointwise Annotation}\label{appendix:pointwise-annotation}
We present the prompts we used to annotate the datasets in Figure~\ref{fig:pretraining-prompt}. We use a similar rubric proposed in \cite{wettig2024quratingselectinghighqualitydata}. 

For each document, a LLM first reasons about the document then outputs a 1-2 sentence natural language critique across each of the five dimensions: writing style, expertise, educational value, fact density/accuracy, and efficiency and an integer score in the range $[1, 5]$, where higher values indicate better quality.

In addition to per-dimension scores, the LLM produces:
\begin{itemize}
    \item An \textbf{Overall score} (an integer in $[1, 5]$)
    \item A \textbf{natural language critique}, consisting of 1-2 sentences summarizing the document's strengths and weaknesses
\end{itemize}

In practice, we rely on the LLM to follow the requested output format. LLMs reliably produced well-structured outputs with valid integer scores in the vast majority of cases. We resample up to 3 times and discard samples that do not comply with the schema after retrying. The overall score is generated after the model generates the other dimension's scores effectively learning a latent aggregation function.

Our annotation pipeline relies on a LLM annotator, as a result:
\begin{itemize}
    \item Scores may be inconsistently calibrated across documents or domains. For this work we did not include domain specific features that should be areas of focus for annotations.
    \item Some dimensions may exhibit correlation or redundancy (e.g., Expertise and Fact Density).
    \item The free-form critiques can vary in style and specificity.
\end{itemize}

We note that the current rubric was designed for general-purpose text quality assessment. Dimensions such as writing style and educational value may not optimally capture quality for domains such as coding, where correctness, modularity, and algorithmic efficiency are more relevant signals. The non-monotonic relationship between quality tier and performance observed for code in RQ1 may partially reflect this mismatch. Our generic rubric likely conflates conventionality with quality, where boilerplate, well-structured code was rated more highly. It's possible that domain-adapted rubrics could improve data separability and learning signal. However, it's also possible that the broadness of the current rubric is sufficient because the model learns to exploit the signal captured by the labeling, even if it does not perfectly align with domain-specific notions of quality.

We present a few example documents and critiques in Figures~\ref{fig:example-critique} ~\ref{fig:dog-cat-statistics-critique}, and~\ref{fig:geometry-logic-critique}.

\newpage

\begin{figure}[H]
\begin{tcolorbox}[title=Annotation Prompt,colback=gray!5,colframe=gray!75!black]
\textbf{Instructions.} Please act as an expert in providing feedback for pre-training documents. You will be given a text snippet. Your job is to think step by step and provide a thoughtful, detailed analysis of its quality based on the rubrics below. Then give a rating for each dimension, followed by a final overall rating (1 to 5).

\medskip
\textbf{Evaluation Dimensions.} Evaluate whether the text demonstrates \textbf{Polished Writing Style, Expertise, Educational Value, Fact Density/Accuracy,} and \textbf{Efficiency}.

\medskip
\textbf{Writing Style.} A good text should exhibit a polished, fluent, and professional writing style that enhances readability and engagement.
\begin{itemize}
    \item \textbf{Fluency \& Polish:} Smooth, grammatically correct, and stylistically refined language.
    \item \textbf{Engagement:} Maintains reader interest through varied sentence structure and appropriate tone.
\end{itemize}

\textbf{Expertise.} A good text should reflect deep domain knowledge and require prerequisite understanding.
\begin{itemize}
    \item \textbf{Depth of Knowledge:} Advanced concepts assuming familiarity with domain terminology.
    \item \textbf{Prerequisite Requirement:} Includes non-trivial insights or complex ideas.
\end{itemize}

\textbf{Educational Value.} A good text should teach effectively through structured explanations.
\begin{itemize}
    \item \textbf{Clarity of Explanations:} Clear reasoning, often step-by-step or with examples.
    \item \textbf{Pedagogical Elements:} Includes questions, comparisons, or summaries that aid learning.
\end{itemize}

\textbf{Fact Density / Accuracy.} A good text should be rich in precise, accurate facts.
\begin{itemize}
    \item \textbf{Fact Density:} High number of non-trivial, verifiable facts.
    \item \textbf{Specificity \& Obscurity:} Emphasizes precise, domain-specific insights.
    \item \textbf{Accuracy:} All claims are correct and grounded in reality.
\end{itemize}

\textbf{Efficiency.} A good text should communicate ideas concisely.
\begin{itemize}
    \item \textbf{Relevance:} Every sentence contributes meaningfully.
    \item \textbf{Conciseness:} Avoids repetition and unnecessary verbosity.
\end{itemize}

\medskip
\textbf{Input.} \texttt{\{text\}}

\medskip
\textbf{Output Format.}
\begin{verbatim}
{
    "Writing Style": {"score": 1-5, "explanation": "..."},
    "Expertise": {"score": 1-5, "explanation": "..."},
    "Educational Value": {"score": 1-5, "explanation": "..."},
    "Fact Density / Accuracy": {"score": 1-5, "explanation": "..."},
    "Efficiency": {"score": 1-5, "explanation": "..."},
    "Overall": {"score": 1-5, "explanation": "..."}
}
\end{verbatim}
\end{tcolorbox}

\caption{Prompt used for pointwise annotation.}
\label{fig:pretraining-prompt}
\end{figure}

\input{Appendix/A_1_example_documents_and_critiques}

\subsection{Pairwise Annotation} \label{appendix:pairwise-annotation}
In addition to the pointwise annotation pipeline described in Appendix~\ref{appendix:pointwise-annotation}, we develop a pairwise annotation procedure that is better suited to regimes where the data distribution is narrow and pointwise scores compress into a small distribution, such as heavily curated SFT mixtures. Rather than asking the LLM judge to score documents independently against a rubric, we ask it to directly compare two documents and indicate which would be more useful for downstream training. 

\textbf{Procedure.} Given a corpus $\mathcal{D}$ of $N$ documents, we sample pairs $(d_i, d_j)$ such that each document participates in $k = 7$ comparisons. The choice of $k$ trades off coverage of the comparisons against annotation compute. We found $k=7$ to be sufficient for the Bradley-Terry fit to converge to a stable ranking on our SFT mixtures. For each pair, we randomize the order in which $d_i$ and $d_j$ are presented to the judge to mitigate position bias, and prompt the judge to output a winner (\texttt{A}, \texttt{B}, or \texttt{tie}) along with a short justification. The full prompt is shown in Figure~\ref{fig:pairwise-prompt}. We use the same judge model as in the pointwise pipeline (Qwen3-30B-A3B~\cite{yang2025qwen3technicalreport}).

\textbf{Bradley-Terry aggregation.} We aggregate the resulting pairwise preferences into a per-document score by fitting a Bradley-Terry model~\cite{bradley-terry}. Each document $d_i$ is assigned a latent strength parameter $\gamma_i \in \mathbb{R}$, and the probability that $d_i$ is preferred over $d_j$ is modeled as
\[
P(d_i \succ d_j) = \frac{\exp(\gamma_i)}{\exp(\gamma_i) + \exp(\gamma_j)}.
\]
Ties are split symmetrically, contributing half a win to each side. We fit $\boldsymbol{\gamma} = (\gamma_1, \dots, \gamma_N)$ by minimizing the negative log-likelihood with L-BFGS~\cite{Liu1989LBFGS}, pinning $\gamma_1 = 0$ during optimization for identifiability and then re-centering the fitted parameters to have mean zero.

\textbf{Bucketing.} The fitted strengths $\boldsymbol{\gamma}$ produce a continuous score over documents. To plug into the \methodname objective without changing the training procedure, we discretize $\boldsymbol{\gamma}$ into the same five-level quality labels used by $\phi_{\text{tok}}$, using percentile cutoffs at the 10th, 30th, 60th, and 85th percentiles of the score distribution. We did not tune bucket cutoffs against downstream performance.

\begin{center}
\begin{minipage}{\linewidth}
\begin{tcolorbox}[
  title=Pairwise Annotation Prompt,
  colback=gray!5,
  colframe=gray!75!black,
  breakable
]
You are an expert evaluator of training data for reasoning models. You will be shown two training examples, each consisting of a question and a step-by-step reasoning response. Your task is to determine which example would be MORE VALUABLE as a training example for teaching a language model to reason well.
\medskip

\textbf{EVALUATION CRITERIA} --- For EACH example, you must carefully analyze:
\begin{enumerate}
    \item \textbf{Correctness:} Is the final answer correct? Verify the answer --- do not assume it is correct just because the reasoning looks plausible. Check for calculation errors, off-by-one mistakes, wrong formulas, or conclusions that don't follow. An incorrect answer makes the example actively harmful for training.
    \item \textbf{Reasoning quality:} Is the reasoning logically sound, well-structured, and free of hallucinated facts or theorems? Does each step follow from the previous one? Does the trace demonstrate good problem-solving habits that a student model should learn? Self-correction is a positive signal; circular reasoning and repeated failed approaches are negative.
    \item \textbf{Efficiency:} Is the reasoning proportionate to the problem's complexity? A concise, direct solution is better than a verbose, meandering one. If the trace is 5x longer than what an expert would write, that is a significant negative even if the answer is correct. Over-explanation of trivial steps and restating the problem multiple times are signs of poor efficiency.
    \item \textbf{Problem difficulty:} Correct solutions to harder problems are more valuable for training than correct solutions to easy problems. A competition-level math problem solved well is worth more than a basic arithmetic problem solved well.
\end{enumerate}

\medskip
\textbf{Example A}
\begin{verbatim}
<example_a>
{sample_a}
</example_a>
\end{verbatim}

\textbf{Example B}
\begin{verbatim}
<example_b>
{sample_b}
</example_b>
\end{verbatim}

\medskip
\textbf{INSTRUCTIONS:} You MUST think extensively before producing your answer. In your thinking, perform the following analysis for EACH example separately:
\begin{enumerate}
    \item What is the problem asking? How difficult is it?
    \item Is the final answer correct? Verify it.
    \item Is the reasoning logically sound? Are there errors, hallucinations, or gaps?
    \item Is the reasoning efficient? How does the trace length compare to what an expert would write?
    \item Overall, how valuable is this example for training?
\end{enumerate}
After analyzing BOTH examples in detail, compare them and make your decision. Your thinking should be thorough --- at least several paragraphs of analysis. Do not rush to a conclusion.
\medskip

After your thinking, respond with this JSON:
\begin{verbatim}
{"winner": "A" or "B" or "tie",
 "reasoning": "1-2 sentence summary of why"}
\end{verbatim}
\end{tcolorbox}

\captionof{figure}{Prompt used for pairwise annotation of SFT examples.}
\label{fig:pairwise-prompt}
\end{minipage}
\end{center}

%% file: Appendix/A_1_example_documents_and_critiques.tex
\begin{tcolorbox}[
  title=Example Annotation Critique,
  colback=gray!5,
  colframe=gray!75!black,
  breakable
]

\textbf{Document:}

\begin{quote}\small
The expression to simplify is likely the fraction:

\[
\frac{c(2a + 6)}{\text{denominator}}
\]

\textbf{Step-by-Step Explanation:}

1. \textbf{Factor the numerator:}
\[
2a + 6 = 2(a + 3)
\]
So the numerator becomes:
\[
c \cdot 2(a + 3) = 2c(a + 3)
\]

2. \textbf{Simplify the fraction:}

If the denominator is also \(2c(a + 3)\), the fraction simplifies to:
\[
\frac{2c(a + 3)}{2c(a + 3)} = 1
\]

If the denominator is only \(2c\), the simplified form is:
\[
\frac{2c(a + 3)}{2c} = a + 3
\]

\textbf{Common Mistake:}

If someone incorrectly factors \(2a + 6\) as \(a + 6\), they might write:
\[
\frac{c(a + 6)}{\text{denominator}}
\]
This is wrong because \(2a + 6 = 2(a + 3)\), not \(a + 6\). Always factor out the greatest common factor first.

\textbf{Final Answer:}

Depends on the denominator. If the denominator is \(2c(a + 3)\), the result is \(1\). If the denominator is \(2c\), the result is \(a + 3\). Clarify the denominator to proceed.
\end{quote}

\medskip

\textbf{Critique:}

\begin{tabular}{p{0.22\linewidth}p{0.08\linewidth}p{0.62\linewidth}}
\textbf{Criterion} & \textbf{Score} & \textbf{Explanation} \\
\hline
Writing Style & 4 & Fluent and grammatically sound with clear mathematical notation, but the opening phrase ``likely the fraction'' introduces unnecessary uncertainty. \\
Expertise & 3 & Assumes basic algebra knowledge but lacks depth by not specifying the denominator's role in advanced contexts. \\
Educational Value & 5 & Strong step-by-step reasoning, explicit common-mistake analysis, and clear conditional outcomes. \\
Fact Density / Accuracy & 5 & The mathematical facts are accurate and the examples are specific. \\
Efficiency & 4 & Concise overall, though the placeholder ``denominator'' reduces precision. \\
Overall & 4 & Strong educational value and accuracy offset minor polish and precision issues. \\
\end{tabular}

\end{tcolorbox}

\captionof{figure}{Example of a document from Crane Math and the associated pointwise critique.}
\label{fig:example-critique}

\newpage

\begin{tcolorbox}[
  title=Example Annotation Critique,
  colback=gray!5,
  colframe=gray!75!black,
  breakable
]

\textbf{Document:}

\begin{quote}\small
\textbf{Question:}

Are dogs heavier than cats? How would you statistically determine this?

\medskip
\textbf{Answer:}

To determine if dogs are heavier than cats, we use \textbf{inferential statistics} to compare the average weights of the two populations. Here's the step-by-step process:

\begin{enumerate}
    \item \textbf{Formulate Hypotheses:}
    \begin{itemize}
        \item \textbf{Null Hypothesis (\(H_0\)):} Dogs and cats have equal average weights.
        \item \textbf{Alternative Hypothesis (\(H_1\)):} Dogs have a significantly higher average weight than cats.
    \end{itemize}

    \item \textbf{Sampling:}
    \begin{itemize}
        \item Collect a \textbf{representative sample} of dog and cat weights using methods like \textbf{stratified sampling}, to account for breeds and genders, or \textbf{simple random sampling}.
        \item Ensure the sample size is sufficient, for example, \(>30\) per group for a t-test.
    \end{itemize}

    \item \textbf{Descriptive Statistics:}
    \begin{itemize}
        \item Calculate \textbf{means} and \textbf{standard deviations} for both groups.
        \item Example:
        \begin{itemize}
            \item Dogs: Mean \(= 30\) lbs, SD \(= 10\) lbs.
            \item Cats: Mean \(= 10\) lbs, SD \(= 3\) lbs.
        \end{itemize}
    \end{itemize}

    \item \textbf{Statistical Test:}
    \begin{itemize}
        \item Use a \textbf{t-test}, for normally distributed data with \(\geq 30\) samples, or a \textbf{Mann--Whitney U test}, for non-normal data or small samples, to compare the groups.
        \item The test calculates a \textbf{p-value}, which indicates the probability of observing the difference if the null hypothesis were true.
    \end{itemize}

    \item \textbf{Interpret Results:}
    \begin{itemize}
        \item If the \textbf{p-value} is less than the significance level, for example \(\alpha = 0.05\), reject the null hypothesis.
        \item Example: If \(p = 0.001\), there is strong evidence that dogs are heavier than cats.
    \end{itemize}

    \item \textbf{Considerations:}
    \begin{itemize}
        \item \textbf{Biased samples}, such as only large dog breeds, can skew results.
        \item \textbf{Variability} in weights, for example dog breeds ranging from 3--180 lbs versus cats ranging from 5--15 lbs, affects statistical power.
    \end{itemize}
\end{enumerate}

\textbf{Conclusion:}

Statistical analysis typically shows dogs are heavier than cats on average. However, the exact result depends on the sample data and proper application of hypothesis testing. For instance, a t-test might reveal a statistically significant difference, such as \(p < 0.05\), confirming the alternative hypothesis.
\end{quote}

\medskip

\textbf{Critique:}

\begin{tabular}{p{0.22\linewidth}p{0.08\linewidth}p{0.62\linewidth}}
\textbf{Criterion} & \textbf{Score} & \textbf{Explanation} \\
\hline
Writing Style & 5 &
Flawless fluency with precise technical terms, varied sentence structure, and professional tone that maintains engagement without jargon overload. \\

Expertise & 5 &
Demonstrates advanced statistical knowledge through nuanced methodological choices, including stratified sampling and the Mann--Whitney U test, as well as domain-specific considerations like breed variability ranges. \\

Educational Value & 5 &
Exceptional pedagogy with step-by-step reasoning, concrete examples, such as 30 lbs mean dog weight, and a critical ``Considerations'' section that anticipates learner misconceptions. \\

Fact Density / Accuracy & 5 &
High-density factual precision: specific weight ranges, such as 3--180 lbs for dogs, exact statistical thresholds, such as \(p < 0.05\), and correct application of inferential methods with no generalizations. \\

Efficiency & 5 &
Zero redundancy; every sentence advances the statistical methodology with no filler, maintaining optimal information density per word. \\

Overall & 5 &
A masterclass in technical education: perfectly balanced expertise, pedagogy, and precision that would serve advanced learners without overwhelming them. \\
\end{tabular}

\end{tcolorbox}

\captionof{figure}{Example of a document from Crane Math and the associated pointwise critique.}
\label{fig:dog-cat-statistics-critique}

\newpage
\begin{tcolorbox}[
  title=Example Annotation Critique,
  colback=gray!5,
  colframe=gray!75!black,
  breakable
]

\textbf{Document:}

\begin{quote}\small
\textbf{Geometry}

\medskip
\textbf{Book edition:} Student Edition

\textbf{Author(s):} Ray C. Jurgensen, Richard G. Brown, John W. Jurgensen

\textbf{ISBN:} 9780395977279

\medskip
\hrule
\medskip

\textbf{Statement Analysis}

\medskip
\textbf{Original Statement:}

\emph{If I live in Los Angeles, then I live in California.}

\textbf{Truth Value:} \textbf{True}

\emph{Los Angeles is a city in California, so anyone living there must live in California.}

\medskip
\hrule
\medskip

\textbf{Contrapositive:}

\emph{If I do not live in California, then I do not live in Los Angeles.}

\textbf{Truth Value:} \textbf{True}

\emph{The contrapositive of a conditional statement is logically equivalent to the original. Since Los Angeles is in California, not living in California implies not living in Los Angeles.}

\medskip
\hrule
\medskip

\textbf{Converse:}

\emph{If I live in California, then I live in Los Angeles.}

\textbf{Truth Value:} \textbf{False}

\emph{California contains many cities, such as San Francisco and Sacramento. Living in California does not guarantee living in Los Angeles.}

\medskip
\hrule
\medskip

\textbf{Inverse:}

\emph{If I do not live in Los Angeles, then I do not live in California.}

\textbf{Truth Value:} \textbf{False}

\emph{People can live in other cities in California, such as San Diego, so not living in Los Angeles does not imply not living in California.}

\medskip
\hrule
\medskip

\textbf{Completed Table:}

\medskip
\begin{tabular}{p{0.22\linewidth}p{0.58\linewidth}p{0.12\linewidth}}
\textbf{Statement Type} & \textbf{Statement} & \textbf{True/False} \\
\hline
1. Given &
If I live in Los Angeles, then I live in California. &
True \\

2. Contrapositive &
If I do not live in California, then I do not live in Los Angeles. &
True \\

3. Converse &
If I live in California, then I live in Los Angeles. &
False \\

4. Inverse &
If I do not live in Los Angeles, then I do not live in California. &
False \\
\end{tabular}

\medskip
\hrule
\medskip

\textbf{Key Concepts:}

\begin{itemize}
    \item \textbf{Contrapositive} is logically equivalent to the original statement.
    \item \textbf{Converse} and \textbf{Inverse} are not logically equivalent to the original and must be evaluated separately.
    \item Truth values depend on real-world context, such as geographic relationships.
\end{itemize}
\end{quote}

\medskip

\textbf{Critique:}

\begin{tabular}{p{0.22\linewidth}p{0.08\linewidth}p{0.62\linewidth}}
\textbf{Criterion} & \textbf{Score} & \textbf{Explanation} \\
\hline
Writing Style & 3 &
The writing is grammatically correct and clear but lacks stylistic sophistication or engagement. The formulaic structure, including bulleted statements and a table, prioritizes clarity over fluid prose, resulting in a dry, textbook-like tone without varied sentence structure or rhetorical flair. \\

Expertise & 2 &
The content presents foundational logic concepts, such as contrapositive and converse, at a basic level, requiring no domain-specific prerequisites. It uses trivial real-world examples, namely the geography of California, rather than advanced mathematical insights, failing to demonstrate depth or challenge readers. \\

Educational Value & 4 &
Effectively teaches core logical relationships through structured examples, clear truth-value explanations, and a well-organized table. The inclusion of real-world context, such as San Francisco and Sacramento, and key concept summaries enhances pedagogical clarity for beginners. \\

Fact Density / Accuracy & 3 &
Facts are accurate but lack density and obscurity. Geographic relationships, such as Los Angeles being in California, are common knowledge rather than domain-specific trivia. The text relies on widely known examples rather than verifiable niche data or nuanced details. \\

Efficiency & 5 &
Every element serves a purpose: the table condenses key information, explanations avoid redundancy, and examples directly illustrate concepts. No filler content exists, making the text maximally concise for its educational goal. \\

Overall & 3 &
While highly efficient and pedagogically sound for introductory geometry, the text lacks advanced expertise, obscure facts, and stylistic engagement. It successfully teaches basics but falls short of the rubric's expectations for depth and intellectual challenge. \\
\end{tabular}

\end{tcolorbox}

\captionof{figure}{Example rubric-based critique for a geometry training-data annotation, shown with the evaluated document.}
\label{fig:geometry-logic-critique}

%% file: Appendix/method_details.tex
\newpage

\section{Methodology Details}\label{appendix:methodology-details}
\subsection{Training Details} \label{appendix:training-details}
\textbf{Pretraining and Continued Pretraining.} For the 7.5B transformer, we follow the optimal hyperparameter calculations of \cite{deepseekai2024deepseekllmscalingopensource} assuming we were going to train the model for a total of 125B tokens. This yields a peak learning rate of $6.1e-4$ and a batch size of 448. We use AdamW \cite{loshchilov2019decoupledweightdecayregularization} with $(\beta_1=0.9, \beta_2=0.95)$ with a weight decay of 0.1 and gradient clipping of 1.0. We adopt a WSD schedule with 2285 steps of warmup and a decay phase that decays the lr to $6.1\times10^{-6}$. For CPT, we precompute the total token budget and begin the decay phase to cover the last 15\% of tokens. We use a context length of 8192 for pretraining and continued pretraining. \\

For the 12B Mamba-Transformer hybrid model, we initialize from the 12T and 18T token checkpoints for up to 34B tokens. Since the CPT horizon is short relative to pretraining, we resume the original LR schedule from the original LR schedule and decay to 1\% of the peak learning rate to $4.5e-6$. We also inherit the batch size of 736 and context length of 8192 from pretraining. \\

\textbf{7.5B Transformer Architecture.} The 7.5B model is a decoder-only transformer with 30 layers, a hidden size of 5120, and a FFN hidden size of 13824. We used group-query attention \cite{ainslie2023gqatraininggeneralizedmultiquery} with 40 query heads and 8 KV groups (with a head dimension of 128). Positional information is provided via RoPE with base $10^6$ applied to the full head dimension. We use squared-ReLU \cite{so2022primersearchingefficienttransformers} activations in the FFN, RMSNorm for layer normalization, and untie the input embedding and output projection. All linear layers are bias-free. We do not use dropout. \\

\textbf{Supervised fine-tuning.} We perform SFT on the 7.5B transformer. We performed an initial learning rate sweep over $lr=(1e-6, 2e-6, 5e-6, 1e-5)$ on the OpenThinker dataset 
and identified $1e-5$ as the optimal learning rate, which we used across all subsequent runs. For all SFT datasets we trained for 1.2M samples, a batch size of 512, maximum sequence length of 32768, and warmup over the first 10\% of training steps.

For SFT we also consider three different datasets:
\begin{itemize}
    \item \textit{General SFT}: We utilize 10 million samples from the Nemotron-Pretraining-SFT-v1 
    \cite{nvidia2025nvidianemotronnano2} general-purpose subset and convert them into SFT samples. We then pack these sequences to the maximum sequence length yielding 168,350 samples.
    \item \textit{OpenThinker}: We clean OpenThinker \cite{guha2025openthoughtsdatarecipesreasoning} to include only complete SFT samples. This yields a total of 392,458 samples. 
    \item \textit{SFT > 4k}: We filter the samples from the Nemotron-Pretraining-SFT-v1~\cite{nvidia2025nvidianemotronnano2} dataset to those longer than $4k$, randomly select 2.2 million samples, and pack the sequences to the maximum sequence length yielding 611,243 samples.
\end{itemize}

\subsection{Evaluation Details} \label{appendix:eval-details}
\textbf{Pretraining evaluation details.} Using LM Evaluation Harness \cite{Leo_Gao_Jonathan}, we conduct a wide range of evaluations covering:

\begin{itemize} 
\item \textbf{Math}. We evaluate the mathematical reasoning ability of models in a zero-shot manner with two key math benchmarks GSM8k \cite{cobbe2021trainingverifierssolvemath} and MATH \cite{hendrycks2021measuringmathematicalproblemsolving}.
\item \textbf{Coding}. We test the zero-shot coding ability of models with two coding benchmarks HumanEval \cite{chen2021evaluatinglargelanguagemodels}, MBPP \cite{austin2021programsynthesislargelanguage}. For both we use the EvalPlus variants and sanitization of generations \cite{liu2023codegeneratedchatgptreally}. We evaluate pass@1 averaged over $n=20$ generations per prompt. For MBPP, we modified the evaluation to include the reference answer's function name in the answer, see Appendix~\ref{appendix:mbpp-drop} for more details.
\item \textbf{General Purpose}. We consider 4 commonsense and logical reasoning evaluations in 0-shot settings: ARC Challenge \cite{clark2018thinksolvedquestionanswering}, RACE \cite{LaiRACE2017}, MMLU \cite{hendrycks2021measuringmassivemultitasklanguage}, and MMLU-Pro \cite{wang2024mmluprorobustchallengingmultitask}.
\end{itemize}

\textbf{SFT evaluation details.}
For SFT, we use Nemo-Skills\footnote{\url{https://github.com/NVIDIA-NeMo/Skills}}. We cover a wide range of evaluations including:
\begin{itemize}
    \item \textbf{Math.} We evaluate the ability of models to answer math questions in a zero-shot setting. We average scores across AMC23 (15 repeats) \cite{AMC}, AIME 24/25 (15 repeats) \cite{AIME2024, AIME2025}, GSM8k (1 repeat) \cite{cobbe2021trainingverifierssolvemath}, MATH-500 (4 repeats) \cite{hendrycks2021measuringmathematicalproblemsolving}, Minerva (4 repeats) \cite{lewkowycz2022solvingquantitativereasoningproblems}, and OlympiadBench (4 repeats) \cite{he2024olympiadbenchchallengingbenchmarkpromoting}.
    \item \textbf{Coding.} We evaluate the ability of models to answer coding questions with HumanEval (20 repeats) \cite{chen2021evaluatinglargelanguagemodels} and LiveCodeBench (4 repeats) \cite{jain2024livecodebenchholisticcontaminationfree}.
    \item \textbf{General.} We evaluate other general capabilities via MMLU (1 repeat) \cite{hendrycks2021measuringmassivemultitasklanguage}, MMLU-Pro (1 repeat) \cite{wang2024mmluprorobustchallengingmultitask}, and IFEval (1 repeat) \cite{zhou2023instruction}.
\end{itemize}
For evaluations with multiple repeats, we sample with temperature $= 0.6$, $top\_k = -1$, and $top\_p = 0.95$, and report the average score across repeats. For single-repeat evaluations, we use greedy decoding.\\

\begin{longtable}{c c p{0.7\textwidth}}
\caption{Full set of critique prefixes used for evaluation, organized by domain and score tier. Score ``--'' denotes short/generic variants.}
\label{tab:critique_prefixes} \\
\toprule
\textbf{Domain} & \textbf{Score} & \textbf{Critique Prefix} \\
\midrule
\endfirsthead
\toprule
\textbf{Domain} & \textbf{Score} & \textbf{Critique Prefix} \\
\midrule
\endhead
\midrule
\multicolumn{3}{r}{\textit{Continued on next page}} \\
\bottomrule
\endfoot
\bottomrule
\endlastfoot

\multicolumn{3}{l}{\textit{Math Domain}} \\
\midrule
Math & 5 & A near-perfect example of technical educational content: accurate, efficient, deeply knowledgeable, and pedagogically sound, with only
minor room for improvement. \\
Math & 5 & A model technical document exhibiting near-perfect alignment with all rubric criteria: expert-level content, flawless execution, and
maximal information density. \\
Math & 5 & A near-perfect synthesis of all rubric dimensions: professionally written, expertly accurate, pedagogically rigorous, fact-rich, and
maximally concise. \\
Math & 5 & A masterclass in technical documentation: perfectly balanced expertise, flawless execution of educational principles, and maximum
information density. Every element serves the core objective with surgical precision. \\
Math & 5 & A masterclass in technical communication---perfectly balanced for its educational purpose, with no room for improvement in clarity, depth,
or utility. \\
Math & 5 & Exceptional technical document that excels in all dimensions. The precision, depth, and clarity make it ideal for pre-training, with only
minor room for pedagogical enhancement. \\
Math & 5 & A model pre-training document: flawless accuracy, deep expertise, and exceptional educational structure. The minor writing style limitation
is outweighed by its unmatched technical precision and pedagogical rigor. \\
\midrule
Math & 4 & A well-crafted, pedagogically strong introductory document that excels in clarity and structure but remains firmly at a basic level without
deeper domain insights or unique factual density. \\
Math & 4 & Exceptionally accurate and efficient with strong expertise, but educational value is hampered by the absence of explanatory depth or
interactive learning elements. \\
Math & 4 & Exceptional technical accuracy and efficiency with strong expertise, but limited educational engagement reduces the overall impact for
learning purposes. \\
Math & 4 & Strong practical utility with excellent efficiency and clear educational value, but limited by moderate expertise depth and modest fact
density beyond basic math concepts. \\
Math & 4 & Strong technical execution and efficiency with clear educational value, but limited by shallow expertise and negligible fact density beyond
basic curriculum material. \\
\midrule
Math & 3 & Adequate coverage of the topic with acceptable accuracy, but lacking depth, expert insights, and efficient organization that would elevate
it beyond introductory material. \\
Math & 3 & While accurate and efficient, the text fails to deliver meaningful educational depth or expertise. It functions as a basic example but
lacks the sophistication expected for a pre-training document targeting advanced learners. \\
Math & 3 & A technically sound but fundamentally basic resource that meets minimal educational requirements but lacks depth, advanced insights, or
meaningful pedagogical innovation for its audience. \\
Math & 3 & A competent but superficial explanation of a basic concept. It fulfills minimal educational requirements but lacks depth, engagement, or
distinctive value expected of advanced technical material. \\
\midrule
Math & 2 & Below-average content with frequent inaccuracies, poor organization, and minimal educational value. Requires significant revision to meet
basic quality standards. \\
Math & 2 & Despite strong efficiency and writing polish, the text fails to demonstrate expertise or educational depth. Its elementary content renders
it unsuitable for pre-training documents requiring advanced rigor. \\
Math & 2 & This is a technically correct but overly simplistic explanation unsuitable for pre-training material. It fails to demonstrate advanced
knowledge or deliver meaningful educational value beyond elementary math. \\
\midrule
Math & 1 & A fundamentally broken draft that fails all core educational purposes. The trivial content, unprofessional presentation, and critical
errors render it useless for pre-training or learning. \\
Math & 1 & The text is fundamentally mismatched for pre-training documents. It fails to demonstrate expertise or provide meaningful value for advanced
learners. \\
\midrule
Math & -- & High quality content. \\
Math & -- & Excellent, expert-level content. \\
Math & -- & A masterclass in technical documentation. \\
Math & -- & Flawless accuracy and deep expertise. \\
Math & -- & Exceptional in all dimensions. \\
Math & -- & Poor quality content. \\
\midrule
\multicolumn{3}{l}{\textit{Code Domain}} \\
\midrule
Code & 5 & Exceptionally well-crafted for its purpose: accurate, efficient, and pedagogically sound with advanced technical depth, limited only by the
inherent constraints of code documentation. \\
Code & 5 & An exemplary technical document that excels in all dimensions. The code's precision, depth, and practical utility make it a gold standard
for implementation guides, with only minor room for enhanced pedagogical elements. \\
Code & 5 & A model technical document that excels across all dimensions---perfectly balanced between precision, educational utility, and
implementation efficiency for its target audience. \\
Code & 5 & A masterclass in technical implementation: flawless accuracy, optimal efficiency, and expert-level depth that sets the standard for
high-quality pre-training code documents. \\
Code & 5 & Peak implementation quality with exceptional expertise and pedagogical value. Every line of code and documentation serves a clear
educational and functional purpose. \\
\midrule
Code & 4 & Exceptional technical implementation and accuracy, but severely limited educational value due to absence of teaching elements. Strong as
documentation, weak as pedagogical material. \\
Code & 4 & Exceptional technical documentation with strong expertise and fact density, but fails as educational material. Strong for API reference,
weak for teaching. \\
Code & 4 & Exceptional as technical reference code but fundamentally misaligned with educational purpose. A high-quality implementation rather than a
training document. \\
Code & 4 & Expert-level implementation with clean structure and advanced domain knowledge, but lacks explanatory depth needed for pedagogical impact.
\\
\midrule
Code & 3 & A technically sound but fundamentally shallow implementation. While efficient and accurate, it fails to provide meaningful educational
value or demonstrate expertise beyond introductory programming. \\
Code & 3 & Solid technical reference with high accuracy and efficiency, but fails to deliver educational depth or advanced expertise expected in
training materials. \\
Code & 3 & A technically sound but educationally shallow code snippet. While efficient and accurate, it fails to leverage code as a teaching tool,
making it suitable for reference but not for learning. \\
Code & 3 & Competent implementation of a standard pattern but lacks depth, advanced insights, or meaningful pedagogical innovation for its audience.
\\
\midrule
Code & 2 & Strong technical implementation but fundamentally misaligned with educational purpose; it is a functional code snippet, not an educational
document. \\
Code & 2 & Severely mismatched with evaluation criteria. While technically sound, it fails to meet the core expectations for a pre-training document.
\\
Code & 2 & Fundamentally misaligned with expectations. Critical accuracy errors and lack of pedagogical value prevent higher ratings despite concise
structure. \\
\midrule
Code & 1 & The text fundamentally misaligns with the rubric requirements. It is a code snippet, not a written pre-training document, making all
dimensions except efficiency inapplicable. \\
Code & 1 & The snippet fundamentally fails to meet the criteria for a pre-training document, as it lacks all essential elements of written content and
functions only as a technical artifact. \\
\midrule
Code & -- & Exemplary code with exceptional depth and accuracy. \\
Code & -- & Gold standard technical implementation. \\
Code & -- & Expert-level code documentation. \\
Code & -- & Technically sound implementation. \\
Code & -- & Poor quality code with limited value. \\

\end{longtable}

\textbf{Critique evaluation details.} During training, each document was prepended with a natural-language critique annotation in the format "\{explanation\}\textbackslash n\textbackslash n\{document\}", where the explanation is a rubric-aligned quality assessment produced by the judge model. At inference time, we exploit this learned conditioning by prepending a \textit{critique prefix} to every evaluation prompt, steering the models toward generating responses consistent with a particular quality tier. To construct these prefixes, we used a coding agent (Claude)\footnote{https://claude.ai/} to search over the critique annotations present in the training data and create representative prefixes for each score level. We additionally included short generic variants (e.g. "\textit{High quality content.}"). The full set of critique prefixes for both domains is listed in Table~\ref{tab:critique_prefixes}. \\

For each model checkpoint evaluated with critiques, we conducted a systematic search: evaluating all candidate prefixes across our pre-training evaluations. Each prefix was prepended to every prompt at serving time. For domain-specific experiments (e.g. CraneMath, SwallowCode), we searched over only the corresponding domain-specific prefix set (27 math prefixes or 23 code prefixes). For the Dolmino experiment, where the model was trained on critiques spannnig both math and code data, we searched over the full combined set of 50 prefixes to identify the optimal conditioning signal across domains. \\

We conduct both training and evaluation on NVIDIA DGX H100 and DGX GB200 systems. Depending on the setup, we would use 7, 8, 14, or 16 nodes for training.

%% file: Appendix/C_mixture.tex
\newpage
\section{Dolmino Mixture} \label{app:dolmino-mix}

We reweight the default Dolmino mixture~\cite{olmo2025olmo3} to better align with \methodname's goal of learning from quality-conditioned data. Our reweighting principle was to upweight structured domains where quality annotation produces the strongest signal (math, code, and scientific text) and downweight more heterogeneous general-purposes sources where the quality signal is noisier. Table~\ref{tab:dolmino-mix-full} presents the full mixture composition for both the baseline and reweighted blends.

\begin{table}[h]
\centering
\small
\begin{tabular}{lrrrrr}
\toprule
 & & \multicolumn{2}{c}{\textbf{Baseline Dolmino}} & \multicolumn{2}{c}{\textbf{Reweighted (Ours)}} \\
\cmidrule(lr){3-4} \cmidrule(lr){5-6}
\textbf{Source} & \textbf{Tokens} & \textbf{Weight} & \textbf{Epochs} & \textbf{Weight} & \textbf{Epochs} \\
\midrule
stack-exchange              & 10.9B & 10.24\% & 1.00 & 10.40\% & 1.01 \\
OLMOCR Science PDFs (High Q.) & 4.3B  & 4.06\%  & 1.00 & 3.40\%  & 0.84 \\
common-crawl (High Q.)      & 22.8B & 21.46\% & 1.00 & 7.50\%  & 0.35 \\
crane-code                  & 10.5B & 9.87\%  & 1.00 & 10.45\% & 1.06 \\
crane-math                  & 5.9B  & 5.54\%  & 1.00 & 11.70\% & 2.11 \\
megamatt                    & 1.9B  & 1.75\%  & 1.00 & 3.70\%  & 2.11 \\
stem-crawl                  & 5.1B  & 4.82\%  & 1.00 & 7.50\%  & 1.55 \\
math                        & 1.6B  & 1.55\%  & 1.00 & 3.30\%  & 2.13 \\
code                        & 0.6B  & 0.61\%  & 1.00 & 1.30\%  & 2.13 \\
Reddit to Flashcards        & 9.5B  & 8.96\%  & 1.00 & 3.30\%  & 0.37 \\
Dolmino 1 Flan              & 5.0B  & 4.71\%  & 1.00 & 5.05\%  & 1.07 \\
OMR Rewrite FullThoughts    & 0.9B  & 0.82\%  & 1.00 & 0.60\%  & 0.73 \\
Nemotron Synth QA           & 5.1B  & 4.83\%  & 1.00 & 7.50\%  & 1.55 \\
Dolmino Math                & 11.2B & 10.53\% & 1.00 & 10.50\% & 1.00 \\
Wiki to RCQA                & 3.1B  & 2.93\%  & 1.00 & 3.30\%  & 1.12 \\
General Reasoning Mix       & 7.8B  & 7.33\%  & 1.00 & 10.50\% & 1.43 \\
\midrule
\textbf{Total}              & \textbf{106.2B} & \textbf{100.00\%} & & \textbf{100.00\%} & \\
\bottomrule\\
\end{tabular}
\caption{Dolmino mixture composition. \textit{Tokens} is the total size of each source corpus. \textit{Weight} is the proportion of the final training blend drawn from that source. \textit{Epochs} is the effective number of passes through the source under the blend (weight $\times$ total training tokens / source tokens). Our reweighting upweights structured domains (crane-math, crane-code, stem-crawl, math, code) and downweights heterogeneous sources (common-crawl, Reddit to Flashcards).}
\label{tab:dolmino-mix-full}
\end{table}

Under the reweighted blend, $\mathcal{D}_{\text{Crane}} = \mathcal{D}_{\text{CraneMath}} \cup \mathcal{D}_{\text{CraneCode}}$ accounts for approximately 22\% of training tokens. All three training strategies (NTP, All Annotations, Specific Subset Annotation) use the same reweighted mixture; they differ only in which documents receive quality-conditioning prefixes during training.

%% file: Appendix/B_MBPP.tex
\newpage
\section{Analysis of MBPP Score Degradation Under \methodname Prefix Tokens}
\label{appendix:mbpp-drop}

\textbf{Experimental setup}
The analysis in this section is conducted on $\mathcal{M}_{base}$ after we continued training with the Dolmino data mix with \methodname. All variants, \methodname no cond., and the five \methodname quality token levels (\textit{[low],} \textit{[medium-low]}, \textit{[medium]}, \textit{[medium-high]}, \textit{[high]}), share the same checkpoint and only differ in which prefix token is prepended at inference time. Evaluation uses MBPP-sanitized with pass@$k$, where $k=20$ samples, temperature 0.6, top-$p$ 0.95, and 768 max generation tokens.

\begin{table}[h]
\centering
\begin{tabular}{lcccc}
\toprule
\textit{method} & MBPP zero-shot performance\\
\midrule
\methodname w/o cond. & 49.6 \\
\methodname + \textit{[low]} & 25.3\\
\methodname + \textit{[medium-low]} & 13.3 \\
\methodname + \textit{[medium]} & 13.3 \\
\methodname + \textit{[medium-high]} & 12.5\\
\methodname + \textit{[high]} & 21.8 \\
\bottomrule\\
\end{tabular}
\caption{MBPP pass@1 scores across different quality token levels. Scores degrade across all prefix levels when compared to scores w/o conditioning}
\end{table}

\textbf{Primary Cause: Function Name Renaming}
The primary cause of the zero-shot score drop is \textit{function name mismatch}. MBPP evaluation relies on test assertions that call the expected function by its name (e.g. \textit{assert find\_Volume(...)}. When the model generates a function with a different name, even if the implementation logic is correct, the test fails with a \textit{NameError}.

We compared per-task generations between the \methodname w/o cond. and \methodname + \textit{[high]}. \methodname conditioned models produce more descriptive, ``production-style'' function names (e.g., \texttt{find\_Volume} $\rightarrow$ \texttt{calculate\_triangular\_prism\_volume} and \texttt{is\_octagonal} $\rightarrow$ \texttt{calculate\_nth\_octagonal\_number}). We analyze all 20 sampled generations per task. \methodname + \textit{[high]} produces a mismatched function name in $58.3\%$ of the generations on average, and in 60.2\% of the tasks the majority of the generations use the wrong name. Whereas, \methodname w/o cond. produces mismatches only 3.9\% of the time and just 1.6\% of tasks have a majority of mismatched names.

To isolate the effect of the function name mismatch from other factors we partition all \methodname + \textit{[high]} generations by whether the first defined function name matches the expected name and present the results in table~\ref{tab:mbpp-name-impact}.

\begin{table}[h]
\centering
\begin{tabular}{lrr}
\toprule
\textbf{Category} & \textbf{\% Generations} & \textbf{Pass Rate} \\
\midrule
\midrule
\textit{\methodname w/o cond.}\\
Correct name & 90.9\% & 50.3\% \\
Wrong name & 3.9\% & 19.9\% \\
No function defined / no expected name & 5.2\%  & 60.4\% \\
\midrule
\textit{\methodname + \textit{[high]}}\\
Correct name   & 36.6\% & 53.9\% \\
Wrong name     & 58.3\% &  2.4\% \\
No function defined / no expected name & 5.1\% & 14.2\% \\
\bottomrule\\
\end{tabular}
\caption{Pass rates for \methodname w/o cond. and \methodname + \textit{[high]} generations partitioned by function name correctness (MBPP).}
\label{tab:mbpp-name-impact}
\end{table}

Generations with the correct function name pass at 53.9\%; while those with a wrong name only pass at 2.4\% and nearly all of the latter fail with \texttt{NameError}. For comparison, without conditioning the samples with the correct function name pass at 50.3\%, confirming that \methodname + \textit{[high]} is at least as capable when the name matches. This failure mode is an evaluation artifact rather than a true measure of coding ability. The model can produce functionally correct code but fails the name-matching requirement of the test harness. \\

\textbf{MBPP-Signature: A more robust zero-shot evaluation variant.}
To directly address the function-name-mismatch artifact, we introduce \textbf{MBPP-Signature}, a modified version of the MBPP evaluation that includes the function signature in the prompt. Rather than presenting the model with only the natural-language task description and a test example, MBPP-Signature extracts the expected function signature, along with any necessary import preamble from the reference solution, and appends it to the prompt. The model is then tasked with generating only the \textit{function body}, similar to the HumanEval evaluation paradigm. 

Concretely, the prompt format changes from:
\begin{verbatim}
"""
Write a function to find the volume of a triangular prism.
assert find_Volume(10,8,6)==240
"""
\end{verbatim}
to:
\begin{verbatim}
"""
Write a function to find the volume of a triangular prism.
assert find_Volume(10,8,6)==240
"""
def find_Volume(l, b, h):
\end{verbatim}

During evaluation, the generated body function is prepended with the preamble and signature before execution against test cases. This ensures that the function name always matches what the tests expect, eliminating the \texttt{NameError}.

\textbf{Results}
Table~\ref{tab:mbpp-signature-results} shows that MBPP-Signature eliminates the zero-shot score gap between the \methodname models without conditioning and conditioned with \textit{[high]}. With the function name anchored, the higher-quality code generation induced by conditioning with prefix tokens becomes an \emph{advantage}, with the best performing prefix outperforming the non-conditioned model by 9.4\% (61.5\% vs 52.1\%).

\begin{table}[h]
\centering
\begin{tabular}{lcc}
\toprule
\textbf{Level} & \textbf{MBPP} & \textbf{MBPP-Signature} \\
\midrule
\methodname w/o cond. & 49.6 & 52.1 \\
\methodname + \textit{[high]} & 21.8 & 59.4 \\
\methodname + \textit{[medium-high]} & 12.5 & 59.9 \\
\methodname + \textit{[medium]} & 13.3 & 61.3 \\
\methodname + \textit{[medium-low]} & 13.3 & 61.5 \\
\methodname + \textit{[low]} & 25.3 & 61.5 \\
\bottomrule\\
\end{tabular}
\caption{Scores for \methodname with and without conditioning on MBPP and MBPP-Signature. With MBPP signature, \methodname with conditioning is able to recover and surpass the performance of not conditioning}
\label{tab:mbpp-signature-results}
\end{table}

As a result, for all other evaluations referencing MBPP, we are using the MBPP-Signature version.

%% file: Appendix/D_IXT_scores.tex
\newpage
\section{\methodname Templated Token Behavior} \label{appendix:templated-token-behavior}
\subsection{$\mathcal{M}_{base}$ Domain specific results}

\begin{figure*}
\includegraphics[width=\textwidth]{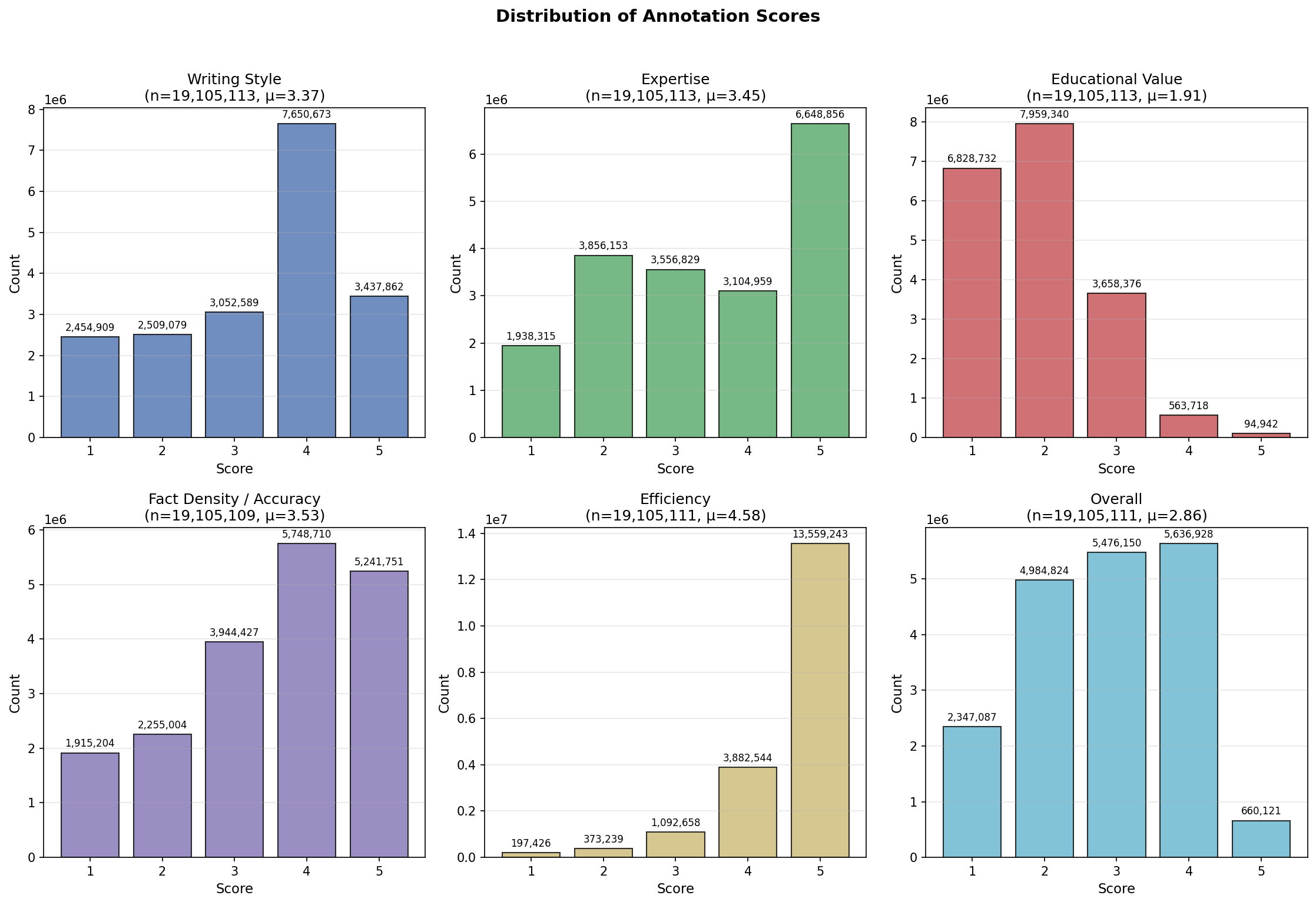}
\caption{Score distribution for Swallowcode with the pointwise rubric. Most of the \textit{overall} scores fall into buckets 1-4 and bucket 5 makes up only \textasciitilde 3.5\% of the total distribution.}
\label{fig:swallowcode-score-dist}
\end{figure*}

Table~\ref{tab:IxT-domain-specific-labels} reports domain specific \methodname results when we continue to train $\mathcal{M}_{base}$ on Crane Math and Swallow Code. Across both domains, conditioning on quality tiers at inference time improves over not conditioning. The best tier leads to \methodname outperforming \ntp by 8.2 points on GSM8k, 10.1 points on MATH, 9.2 points on HumanEval, and 11.0 points on MBPP. The behavior across tiers, however is domain-dependent. On math, performance increases roughly monotonically with tier quality, peaking at \textit{[medium]} for GSM8k and \textit{[medium-high]} for MATH. On code, the relationship is non-monotonic: \textit{[medium-low]} yields the strongest results on both HumanEval and MBPP, while \textit{[high]} underperforms intermediate tiers. We attribute this to the composition of Swallow Code's high tier subset, which constitutes only \textasciitilde 3.5\% of the corpus (see Figure~\ref{fig:swallowcode-score-dist}). This subset skews towards more templated, stylistically narrow examples. \\

We can see how the skewed tiers translates to downstream performance. We classify every failing completion by its execution error type and measure the structural properties of the generated code. Each tier is evaluated at its optimal epoch (the epoch that maximized the tier's performance). \\

\textbf{Code Failure Modes} Table~\ref{tab:failure-modes} reports the error distribution across all 20 sampled completions per problem. The highest tier gains little despite the strongest conditioning signal. On MBPP, using \textit{[high]} as the conditioning signal reduces \texttt{NameError}s only to 4.51\%, while \textit{increasing} the \texttt{ImportError} rate to 11.69\%, nearly canceling out the NameError gains and slightly improving the downstream score by 2.5\%. On HumanEval, conditioning on \textit{[high]} barely reduces the \texttt{NameError}s (14.7\% to 12.2\%) and reduces the final HumanEval score. Conditioning on higher quality tiers appears to bias the model toward importing external libraries unavailable in the execution sandbox, introducing a new failure mode. This is confirmed by the \texttt{ImportError} rate increasing near monotonically with tier.

\begin{table}[t]
\centering
\footnotesize
\begin{tabularx}{\linewidth}{
>{\raggedright\arraybackslash}p{3.2cm}
*{4}{>{\centering\arraybackslash}X}
}
\toprule
& \multicolumn{2}{c}{Math} & \multicolumn{2}{c}{Code} \\
\cmidrule(lr){2-3} \cmidrule(lr){4-5}
Method & GSM8K & MATH & HumanEval & MBPP \\
\midrule
& \multicolumn{2}{c}{\textit{Crane Math}} 
& \multicolumn{2}{c}{\textit{Swallow Code}} \\
\ntp & 43.9 & 38.9 & 35.6 & 49.6 \\
\methodname w/o cond. & 48.2 & 35.2 & 37.5 & 40.1 \\
\methodname + \textit{[high]} & 49.4 & 48.5 & 36.0 & 50.3 \\
\methodname + \textit{[medium-high]} & 49.1 & \textbf{49.0} & 43.2 & 56.6 \\
\methodname + \textit{[medium]} & \textbf{52.1} & 45.1 & 42.0 & 59.6 \\ 
\methodname + \textit{[medium-low]} & 48.5 & 43.5 & \textbf{44.8} & \textbf{60.6} \\
\methodname + \textit{[low]} & 28.7 & 34.4 & 38.5 & 58.5\\
\bottomrule\\
\end{tabularx}
\caption{Domain-specific performance of \methodname on math with Crane Math and code with Swallow Code, where we evaluate the math models on GSM8K and MATH and the coding models on HumanEval and MBPP. We evaluate \ntp, \methodname without any inference-time conditioning, and \methodname+\textit{[tier]}, which indicates inference-time conditioning on the specified quality tier. For math, scores monotonically improve up to intermediate tiers, with \textit{[medium]} and \textit{[medium-high]} yielding the best GSM8K (52.1) and MATH (49.0) results, respectively. For code, the relationship is non-monotonic: intermediate tiers outperform \textit{[high]}, with \textit{[medium-low]} achieving the strongest HumanEval (44.8) and MBPP (60.6).}
\label{tab:IxT-domain-specific-labels}
\end{table}

\begin{table}[h]
\centering
\resizebox{\textwidth}{!}{%
\begin{tabular}{l cccccc r}
\toprule
\textbf{Error Type} & \textbf{no cond.} & \textbf{low} & \textbf{medium-low} & \textbf{med} & \textbf{medium-high} & \textbf{high} & \textbf{$\Delta$(medium-low$-$No-cond.)} \\
\midrule
\multicolumn{8}{l}{\textit{HumanEval}} \\
\midrule
Passed          & 37.5 & 38.6 & \textbf{44.8} & 43.5 & 43.3 & 36.0 & $+$7.3 \\
Wrong answer    & 42.7 & 44.2 & 45.2 & 47.4 & 46.1 & 45.5 & $+$2.6 \\
\texttt{NameError}       & 14.7 & 11.8 & \textbf{3.1}  & 2.6  & 3.2  & 12.2 & $-$11.5 \\
\texttt{ValueError}      & 1.3  & 1.2  & 1.40  & 1.8  & 1.7  & 1.7  & $+$0.1 \\
\texttt{TypeError}       & 1.5  & 1.4  & 1.31  & 1.2  & 1.2  & 1.6  & $-$0.2 \\
\texttt{ImportError}     & 0.3  & 0.8  & 1.89  & 1.1  & 2.3  & 0.8  & $+$1.6 \\
\texttt{IndexError}      & 0.8  & 0.9  & 1.04  & 1.0  & 1.0  & 0.8  & $+$0.2 \\
\midrule
\multicolumn{8}{l}{\textit{MBPP}} \\
\midrule
Passed          & 47.8 & 58.5 & \textbf{60.5} & 59.6 & 56.6 & 50.3 & $+$12.7 \\
Wrong answer    & 29.9 & 31.0 & 31.6 & 30.7 & 30.4 & 28.5 & $+$1.8 \\
\texttt{NameError}       & 12.5 & 1.0  & \textbf{0.8}  & 0.8  & 0.9  & 4.5  & $-$11.8 \\
\texttt{ImportError}     & 5.8  & 6.3  & 4.14  & 5.4  & 8.3  & 11.7 & $-$1.7 \\
\texttt{TypeError}       & 2.6  & 2.2  & 2.26  & 2.5  & 2.5  & 3.2  & $-$0.3 \\
\bottomrule\\
\end{tabular}%
}
\caption{Failure mode distribution (\% of all completions) at each tier's best epoch. Only categories $\geq$1\% in at least one tier are shown.}
\label{tab:failure-modes}
\end{table}

\begin{table}[h]
\centering
\begin{tabular}{l cccccc}
\toprule
\textbf{Metric} & \textbf{no cond.} & \textbf{low} & \textbf{medium-low} & \textbf{medium} & \textbf{medium-high} & \textbf{high} \\
\midrule
\multicolumn{7}{l}{\textit{HumanEval}} \\
\midrule
Avg characters            & 1322  & 1163  & 1245  & 1296  & 1435  & 1284 \\
Avg lines                 & 42.7  & 38.1  & 41.7  & 43.4  & 47.3  & 41.7 \\
Extra \texttt{def} rate   & 0.7  & 0.6  & 0.7  & 0.7  & 0.7  & 0.6 \\
Avg \texttt{for}-loops    & 2.3  & 2.1  & 2.5  & 2.6  & 2.7  & 2.3 \\
Avg \texttt{if}-statements& 4.1  & 3.7  & 4.2  & 4.4  & 4.8  & 3.8 \\
Avg \texttt{try} blocks   & 0.1  & 0.1  & 0.1  & 0.1  & 0.1  & 0.1 \\
\midrule
\multicolumn{7}{l}{\textit{MBPP}} \\
\midrule
Avg characters            & 1429  & 898   & 915   & 1069  & 1234  & 1474 \\
Avg lines                 & 47.6  & 30.3  & 30.8  & 35.8  & 41.0  & 49.1 \\
Extra \texttt{def} rate   & 0.6  & 0.3  & 0.3  & 0.3  & 0.3  & 0.5 \\
Avg \texttt{for}-loops    & 1.6  & 1.3  & 1.3  & 1.4  & 1.6  & 1.8 \\
Avg \texttt{if}-statements& 2.2  & 2.2  & 2.3  & 2.5  & 2.9  & 3.4 \\
Avg \texttt{try} blocks   & 0.3  & 0.1  & 0.1  & 0.1  & 0.2  & 0.3 \\
\bottomrule\\
\end{tabular}
\caption{Code characteristics at each tier's best epoch, averaged across all completions.}
\label{tab:code-characteristics}
\end{table}

\begin{table}[h]
\centering
\begin{tabular}{l cc cc cc}
\toprule
& \multicolumn{2}{c}{\textbf{Avg chars}} & \multicolumn{2}{c}{\textbf{Avg lines}} & \multicolumn{2}{c}{\textbf{Extra \texttt{def} rate}} \\
\cmidrule(lr){2-3} \cmidrule(lr){4-5} \cmidrule(lr){6-7}
\textbf{Tier} & Pass & Fail & Pass & Fail & Pass & Fail \\
\midrule
No cond.     & 1337 & 1513 & 45.6 & 49.4 & 0.6 & 0.5 \\
low         &  769 & 1080 & 26.8 & 35.2 & 0.2 & 0.3 \\
medium-low  &  805 & 1085 & 28.0 & 35.0 & 0.2 & 0.3 \\
medium      &  961 & 1228 & 33.0 & 39.8 & 0.3 & 0.4 \\
medium-high & 1110 & 1395 & 37.7 & 45.2 & 0.3 & 0.4 \\
high        & 1371 & 1579 & 46.3 & 51.9 & 0.5 & 0.5 \\
\bottomrule\\
\end{tabular}
\caption{Code length for passing vs.\ failing completions on MBPP.}
\label{tab:pass-fail-characteristics}
\end{table}

\textbf{Code Characteristics} Table~\ref{tab:code-characteristics} reports structural properties of the generated code. On HumanEval, the code length is relatively stable across tiers. However, on MBPP, code length inreases monotonically with tier: the \textit{medium-low} tier averages 915 characters while the high tier averages 1474. This is driven by far fewer function definitions and try except blocks. 

Table~\ref{tab:pass-fail-characteristics} splits these metrics by pass/fail status for MBPP. For \textit{[medium-low]} the average failing generation length is 1085 characters and 28 lines, while when conditioning on \textit{[high]} the average generation length is 1579 characters and 46.3 lines. This is showing that when conditioning on \textit{[high]}, the model tends to overgenerate more.

\subsection{$\mathcal{M}_{base}$ Dolmino results} 
\begin{table}[t]
\centering
\footnotesize
\setlength{\tabcolsep}{3.5pt}
\renewcommand{\arraystretch}{1.05}

\begin{tabularx}{\linewidth}{
>{\raggedright\arraybackslash}p{3,2cm}
*{9}{>{\centering\arraybackslash}X}
}
\toprule
Method & GSM8K & MATH & \makecell{Human\\Eval} & \makecell{MBPP} & MMLU & MMLU-Pro & ARC-C & RACE & Avg. \\
\midrule \\
\methodname w/o cond. & 30.9 & 37.6 & 37.3 & 52.1 & 49.2 & 25.8 & 66.5 & 74.9 & 46.8 \\
\methodname + \textit{[low]} & 21.2 & 36.3 & 38.4 & 61.5 & 48.7 & 25.7 & 66.2 & 75.1 & 46.6 \\
\methodname + \textit{[medium-low]} & 32.4 & 43.5 & 46.5 & \textbf{61.5} & 49.7 & 26.9 & 67.6 & \textbf{75.3} & 50.4 \\
\methodname + \textit{[medium]} & 39.9 & 49.7 & 46.8 & 61.3 & 50.7 & 26.4 & 67.6 & \textbf{75.3} & 52.2 \\
\methodname + \textit{[medium-high]} & 41.2 & \textbf{51.6} & \textbf{50.5} & 59.9 & \textbf{50.8} & \textbf{27.3} & \textbf{67.7} & 75.1 & \textbf{53.1} \\ 
\methodname + \textit{[high]} & \textbf{43.9} & 46.6 & 42.8 & 59.4 & 51.0 & \textbf{27.3} & 67.6 & 75.2 & 51.7 \\
\bottomrule \\
\end{tabularx}
\caption{Performance when evaluating $\mathcal{M}_{base}$ trained on dolmino with \methodname applied to \textit{specific subsets}. Conditioning on all tokens besides \textit{[low]} improves over not conditioning. The best performing conditioning token was \textit{[medium-high]}.}
\label{tab:dolmino-quality-tiers}
\end{table}

In table~\ref{tab:dolmino-quality-tiers} we present all quality tiers for when we continued training $\mathcal{M}_{base}$ on Dolmino for 1 epoch with \textit{specific subset annotation}. Conditioning on any other tier than \textit{[low]} improves over the unconditioned baseline, with \textit{[medium-high]} yielding the strongest aggregate performance (53.1 on avg., +6.3 over no conditioning). Notably, the optimal tier is task-dependent: \textit{[high]} maximizes GSM8k (43.9), \textit{[medium-high]} is best on MATH (51.6), HumanEval (50.5), MMLU (50.8), and Arc-Challenge (67.7), while \textit{[medium]} and \textit{[medium-low]} attain the highest scores on RACE (75.3) and MBPP (61.5) respectively. The degraded performance under \textit{[low]} conditioning (46.6, on par with no conditioning), indicates that the model has learned to associate the tier tokens with their corresponding document quality, since steering towards low-quality generations recovers near-baseline behavior rather than improving it. Together, these results suggest that the quality tokens function as an effective inference-time control signals.


%% file: Appendix/E_bucketing.tex
\section{Filtering vs Tagging Data with Quality tokens}

\begin{figure*}
\includegraphics[width=\textwidth]{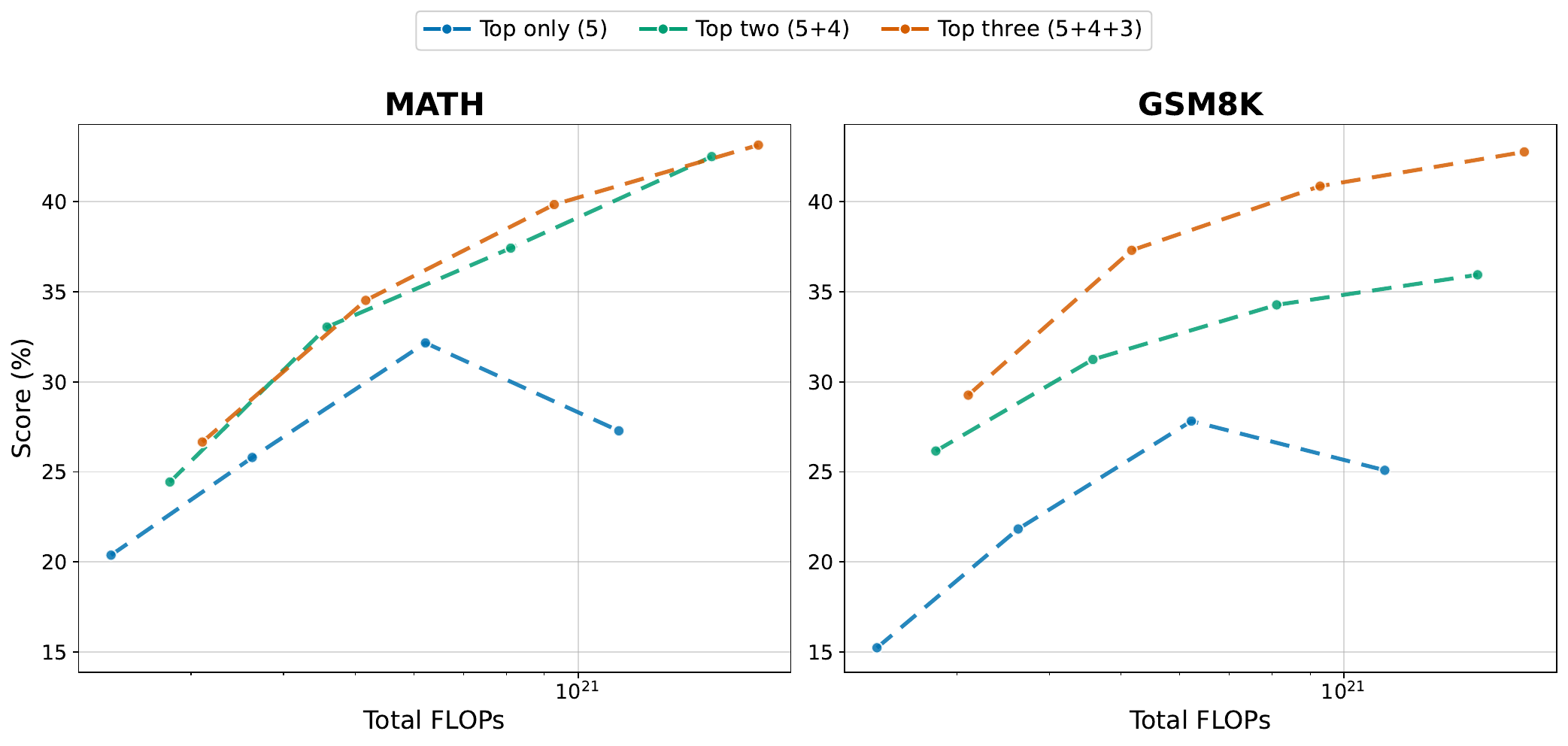}
\caption{FLOP-scaling curves on MATH and GSM8k when continuing to train $\mathcal{M}_{base}$ with \methodname applied to CraneMath. Each curve trains on data annotated with \methodname quality tokens restricted to the top 1, 2, or 3 tiers. Including  lower-quality tiers consistently improves performance across the compute range, with the gap widening at higher FLOP budgets. Notably, training on only the top tier degrades past $10^{21}$ FLOPs, suggesting that quality-filtered subsets alone are insufficient to sustain learning at scale. The results support conditioning on quality tokens over filtering, as the model benefits from seeing diverse data with explicit quality signals. }
\label{Fig:bucket-scaling}
\end{figure*}

A natural question is whether the benefits of quality annotation come from \textit{conditioning} on quality tokens during training, or simply \textit{filtering} low-quality data would suffice. While filtering shown to be effective at \textit{suppressing} undesired capabilities during pretraining~\cite{rathi2026shapingcapabilitiestokenleveldata}, it is less clear whether filtering low-quality/undesirable data when the goal is to \textit{enhance} a target domain. To test this, we train $\mathcal{M}_{base}$ on crane math only using the top quality tier (bucket 5), the top two tiers (buckets 5 + 4), and the top three tiers (buckets 5+4+3), each tagged with the corresponding \methodname quality tokens. Note that all filtered subsets still receive \methodname quality-token prefixes; the comparison thus isolates the effect of retaining lower-quality data with explicit quality signals versus discarding it. Because each filtered subset differs in size, matched epoch counts correspond to different FLOP budgets; Figure~\ref{Fig:bucket-scaling} plots total FLOPs on the x-axis to ensure fair comparison across configurations. For each mix, we train for 1, 2, 4, and 8 epochs. \\

Including lower-quality tiers consistently improves performance across the compute range, with the gap widening at higher FLOP budgets. Most notably, training on only the top tier degrades performance past approximately $10^{21}$ FLOPs on both benchmarks, suggesting that aggressive filtering leads to overfitting when the high-quality subset is exhausted. In contrast, the other configurations improve monotonically. These results indicate that when the goal is to maximize downstream performance rather than suppress capabilities, the model benefits from seeing diverse-quality data with explicit quality signals. These results suggest that, when the goal is to maximize downstream performance on a fixed corpus, conditioning on quality tokens across the full data distribution is preferable to filtering out low-quality subsets alone.

%% file: Appendix/G_SFT_Scores.tex
\section{Detailed SFT Scores}

Table~\ref{tab:fine-grained-scores} reports per-benchmark scores for different SFT training datasets. Benchmarks are organized into three categories: \textit{Math} (AIME24/25, AMC, GSM8k, Math 500, Minerva, OlympiadBench), \textit{Code} (HumanEval, LiveCodeBench), and \textit{General} (MMLU, MMLU-Pro, IFEval).

\begin{table*}[t]
\centering
\vspace{0.5em}
\resizebox{\textwidth}{!}{%
\begin{tabular}{ll ccccccc c cc c ccc c c}
\toprule
& & \multicolumn{8}{c}{\textbf{Math}} & \multicolumn{3}{c}{\textbf{Code}} & \multicolumn{4}{c}{\textbf{General}} & \\
\cmidrule(lr){3-10} \cmidrule(lr){11-13} \cmidrule(lr){14-17}
\textbf{Setting} & \textbf{Method}
& \rotatebox{70}{AIME24} & \rotatebox{70}{AIME25} & \rotatebox{70}{AMC}
& \rotatebox{70}{GSM8k} & \rotatebox{70}{Math 500} & \rotatebox{70}{Minerva}
& \rotatebox{70}{OlympiadBench} & \rotatebox{70}{Avg}
& \rotatebox{70}{HumanEval} & \rotatebox{70}{LiveCodeBench} & \rotatebox{70}{Avg}
& \rotatebox{70}{MMLU} & \rotatebox{70}{MMLU-Pro} & \rotatebox{70}{IFEval} & \rotatebox{70}{Avg}
& \rotatebox{70}{\textbf{Overall Avg.}}
\\
\midrule
\multirow{2}{*}{Openthinker}
& \ntp & 29.3 & \textbf{29.1} & 66.5 & 86.6 & 85.2 & 32.0 & 49.6 & 54.0 & \textbf{70.0} & \textbf{17.9} & \textbf{44.0} & 57.8 & 40.3 & 26.6 & 41.6 & 49.2 \\
& \methodname & \textbf{33.6} & 28.2 & \textbf{69.0} & \textbf{87.0} & \textbf{86.1} & \textbf{34.9} & \textbf{50.0} & \textbf{55.5} & 68.9 & 17.2 & 43.1 & \textbf{58.0} & 40.3 & \textbf{27.0} & \textbf{41.8} & \textbf{50.0} \\
\midrule
\multirow{2}{*}{General SFT}
& \ntp & \textbf{4.7} & 5.1 & 34.8 & 82.2 & 61.4 & \textbf{23.2} & 27.4 & 34.1 & 61.1 & 11.5 & 36.3 & 51.3 & 27.1 & 47.5 & 42.0 & 36.5 \\
& \methodname & 4.0 & \textbf{6.4} & \textbf{36.0} & \textbf{82.3} & \textbf{64.2} & 22.6 & \textbf{28.4} & \textbf{34.9} & \textbf{62.5} & \textbf{12.0} & \textbf{37.2} & \textbf{52.1} & \textbf{28.9} & \textbf{52.7} & \textbf{44.6} & \textbf{37.7} \\
\midrule
\multirow{3}{*}{SFT $>$ 4k}
& \ntp & 28.4 & \textbf{27.1} & 61.5 & 84.7 & 80.1 & 27.0 & 45.1 & 50.6 & 58.1 & \textbf{16.3} & 37.2 & 42.8 & 28.7 & 24.0 & 31.8 & 43.7 \\
& \methodname & 28.4 & 25.3 & \textbf{69.8} & \textbf{85.9} & \textbf{84.7} & 27.0 & \textbf{48.8} & 52.9 & 59.8 & 15.4 & 37.6 & 44.3 & 30.8 & 25.1 & 33.4 & 45.5 \\
& \methodname w/ pairwise ann.  & \textbf{31.8} & 26.7 & 67.0 & 84.9 & 84.4 & \textbf{27.5} & 48.6 & \textbf{53.0} & \textbf{62.3} & 15.3 & \textbf{38.8} & \textbf{45.7} & \textbf{32.9} & \textbf{25.5} & \textbf{34.7} & \textbf{46.1} \\
\bottomrule
\end{tabular}%
}
\caption{Fine-grained benchmark scores for all training configurations on OpenThinker. Benchmarks are grouped into \textbf{Math} (AIME24, AIME25, AMC, GSM8k, Math 500, Minerva Math, OlympiadBench), \textbf{Code} (HumanEval, LiveCodeBench), and \textbf{General} (MMLU, MMLU-Pro, IFEval). Category averages and an overall average across all 12 benchmarks are reported.}
\label{tab:fine-grained-scores}
\end{table*}

%% file: references.bib
@article{hoffmann2022empirical,
  title={An empirical analysis of compute-optimal large language model training},
  author={Hoffmann, Jordan and Borgeaud, Sebastian and Mensch, Arthur and Buchatskaya, Elena and Cai, Trevor and Rutherford, Eliza and de Las Casas, Diego and Hendricks, Lisa Anne and Welbl, Johannes and Clark, Aidan and others},
  journal={Advances in neural information processing systems},
  volume={35},
  pages={30016--30030},
  year={2022}
}

@misc{alpaca,
  author = {Rohan Taori and Ishaan Gulrajani and Tianyi Zhang and Yann Dubois and Xuechen Li and Carlos Guestrin and Percy Liang and Tatsunori B. Hashimoto },
  title = {Stanford Alpaca: An Instruction-following LLaMA model},
  year = {2023},
  publisher = {GitHub},
  journal = {GitHub repository},
  howpublished = {\url{https://github.com/tatsu-lab/stanford_alpaca}},
}

@article{blakeman2025nvidia,
  title={NVIDIA Nemotron 3: Efficient and Open Intelligence},
  author={Blakeman, Aaron and Grattafiori, Aaron and Basant, Aarti and Gupta, Abhibha and Khattar, Abhinav and Renduchintala, Adi and Vavre, Aditya and Shukla, Akanksha and Bercovich, Akhiad and Ficek, Aleksander and others},
  journal={arXiv preprint arXiv:2512.20856},
  year={2025}
}

@article{thrush2024improving,
  title={Improving pretraining data using perplexity correlations},
  author={Thrush, Tristan and Potts, Christopher and Hashimoto, Tatsunori},
  journal={arXiv preprint arXiv:2409.05816},
  year={2024}
}

@inproceedings{shum2025predictive,
  title={Predictive Data Selection: The Data That Predicts Is the Data That Teaches},
  author={Shum, Kashun and Huang, Yuzhen and Zou, Hongjian and Ding, Qi and Liao, Yixuan and Chen, Xiaoxin and Liu, Qian and He, Junxian},
  booktitle={International Conference on Machine Learning},
  pages={55427--55450},
  year={2025},
  organization={PMLR}
}

@inproceedings{penedo2024fineweb,
  title={The FineWeb datasets: decanting the web for the finest text data at scale},
  author={Penedo, Guilherme and Kydl{\'\i}{\v{c}}ek, Hynek and Allal, Loubna Ben and Lozhkov, Anton and Mitchell, Margaret and Raffel, Colin and Von Werra, Leandro and Wolf, Thomas},
  booktitle={Proceedings of the 38th International Conference on Neural Information Processing Systems},
  pages={30811--30849},
  year={2024}
}

@inproceedings{zhuang2025meta,
  title={Meta-rater: A multi-dimensional data selection method for pre-training language models},
  author={Zhuang, Xinlin and Peng, Jiahui and Ma, Ren and Wang, Yinfan and Bai, Tianyi and Wei, Xingjian and Jiantao, Qiu and Zhang, Chi and Qian, Ying and He, Conghui},
  booktitle={Proceedings of the 63rd Annual Meeting of the Association for Computational Linguistics (Volume 1: Long Papers)},
  pages={10856--10896},
  year={2025}
}

@article{mizrahi2025betr,
  title={Language models improve when pretraining data matches target tasks},
  author={Mizrahi, David and Larsen, Anders Boesen Lindbo and Allardice, Jesse and Petryk, Suzie and Gorokhov, Yuri and Li, Jeffrey and Fang, Alex and Gardner, Josh and Gunter, Tom and Dehghan, Afshin},
  journal={arXiv preprint arXiv:2507.12466},
  year={2025}
}

@inproceedings{xie2023doremi,
  title={DoReMi: optimizing data mixtures speeds up language model pretraining},
  author={Xie, Sang Michael and Pham, Hieu and Dong, Xuanyi and Du, Nan and Liu, Hanxiao and Lu, Yifeng and Liang, Percy and Le, Quoc V and Ma, Tengyu and Yu, Adams Wei},
  booktitle={Proceedings of the 37th International Conference on Neural Information Processing Systems},
  pages={69798--69818},
  year={2023}
}

@inproceedings{korbak2023pretraining,
  title={Pretraining language models with human preferences},
  author={Korbak, Tomasz and Shi, Kejian and Chen, Angelica and Bhalerao, Rasika Vinayak and Buckley, Christopher and Phang, Jason and Bowman, Samuel R and Perez, Ethan},
  booktitle={International conference on machine learning},
  pages={17506--17533},
  year={2023},
  organization={PMLR}
}

@article{hatamizadeh2025rlp,
  title={Rlp: Reinforcement as a pretraining objective},
  author={Hatamizadeh, Ali and Akter, Syeda Nahida and Prabhumoye, Shrimai and Kautz, Jan and Patwary, Mostofa and Shoeybi, Mohammad and Catanzaro, Bryan and Choi, Yejin},
  journal={arXiv preprint arXiv:2510.01265},
  year={2025}
}

@article{luo2025language,
  title={Language models can learn from verbal feedback without scalar rewards},
  author={Luo, Renjie and Liu, Zichen and Liu, Xiangyan and Du, Chao and Lin, Min and Chen, Wenhu and Lu, Wei and Pang, Tianyu},
  journal={arXiv preprint arXiv:2509.22638},
  year={2025}
}

@article{chen2024learning,
  title={Learning from natural language feedback},
  author={Chen, Angelica and Scheurer, J{\'e}r{\'e}my and Campos, Jon Ander and Korbak, Tomasz and Chan, Jun Shern and Bowman, Samuel R and Cho, Kyunghyun and Perez, Ethan},
  journal={Transactions on machine learning research},
  year={2024}
}

@article{wang2025text2grad,
  title={Text2Grad: Reinforcement Learning from Natural Language Feedback},
  author={Wang, Hanyang and Wang, Lu and Zhang, Chaoyun and Mao, Tianjun and Qin, Si and Lin, Qingwei and Rajmohan, Saravan and Zhang, Dongmei},
  journal={arXiv preprint arXiv:2505.22338},
  year={2025}
}

@article{chen2021decision,
  title={Decision transformer: Reinforcement learning via sequence modeling},
  author={Chen, Lili and Lu, Kevin and Rajeswaran, Aravind and Lee, Kimin and Grover, Aditya and Laskin, Misha and Abbeel, Pieter and Srinivas, Aravind and Mordatch, Igor},
  journal={Advances in neural information processing systems},
  volume={34},
  pages={15084--15097},
  year={2021}
}

@misc{zhang2025interplay,
      title={On the Interplay of Pre-Training, Mid-Training, and RL on Reasoning Language Models}, 
      author={Charlie Zhang and Graham Neubig and Xiang Yue},
      year={2025},
      eprint={2512.07783},
      archivePrefix={arXiv},
      primaryClass={cs.CL},
      url={https://arxiv.org/abs/2512.07783}, 
}

@misc{xing2025pretrainzero,
      title={PretrainZero: Reinforcement Active Pretraining}, 
      author={Xingrun Xing and Zhiyuan Fan and Jie Lou and Guoqi Li and Jiajun Zhang and Debing Zhang},
      year={2025},
      eprint={2512.03442},
      archivePrefix={arXiv},
      primaryClass={cs.CL},
      url={https://arxiv.org/abs/2512.03442}, 
}

@misc{maini2025safetypretraininggenerationsafe,
      title={Safety Pretraining: Toward the Next Generation of Safe AI}, 
      author={Pratyush Maini and Sachin Goyal and Dylan Sam and Alex Robey and Yash Savani and Yiding Jiang and Andy Zou and Matt Fredrikson and Zacharcy C. Lipton and J. Zico Kolter},
      year={2025},
      eprint={2504.16980},
      archivePrefix={arXiv},
      primaryClass={cs.LG},
      url={https://arxiv.org/abs/2504.16980}, 
}

@article{Akter2025FrontLoadingRT,
  title={Front-Loading Reasoning: The Synergy between Pretraining and Post-Training Data},
  author={Syeda Nahida Akter and Shrimai Prabhumoye and Eric Nyberg and Mostofa Patwary and Mohammad Shoeybi and Yejin Choi and Bryan Catanzaro},
  journal={ArXiv},
  year={2025},
  volume={abs/2510.03264},
  url={https://api.semanticscholar.org/CorpusID:281843895}
}

@inproceedings{Yano2026PretrainingLW,
  title={Pre-training LLM without Learning Rate Decay Enhances Supervised Fine-Tuning},
  author={Yano, Kazuki and Kiyono, Shun and Kobayashi, Sosuke and Takase, Sho and Suzuki, Jun},
  booktitle={The Fourteenth International Conference on Learning Representations},
    year={2026},
}

@misc{olmo2025olmo3,
      title={Olmo 3}, 
      author={Team Olmo and : and Allyson Ettinger and Amanda Bertsch and Bailey Kuehl and David Graham and David Heineman and Dirk Groeneveld and Faeze Brahman and Finbarr Timbers and Hamish Ivison and Jacob Morrison and Jake Poznanski and Kyle Lo and Luca Soldaini and Matt Jordan and Mayee Chen and Michael Noukhovitch and Nathan Lambert and Pete Walsh and Pradeep Dasigi and Robert Berry and Saumya Malik and Saurabh Shah and Scott Geng and Shane Arora and Shashank Gupta and Taira Anderson and Teng Xiao and Tyler Murray and Tyler Romero and Victoria Graf and Akari Asai and Akshita Bhagia and Alexander Wettig and Alisa Liu and Aman Rangapur and Chloe Anastasiades and Costa Huang and Dustin Schwenk and Harsh Trivedi and Ian Magnusson and Jaron Lochner and Jiacheng Liu and Lester James V. Miranda and Maarten Sap and Malia Morgan and Michael Schmitz and Michal Guerquin and Michael Wilson and Regan Huff and Ronan Le Bras and Rui Xin and Rulin Shao and Sam Skjonsberg and Shannon Zejiang Shen and Shuyue Stella Li and Tucker Wilde and Valentina Pyatkin and Will Merrill and Yapei Chang and Yuling Gu and Zhiyuan Zeng and Ashish Sabharwal and Luke Zettlemoyer and Pang Wei Koh and Ali Farhadi and Noah A. Smith and Hannaneh Hajishirzi},
      year={2025},
      eprint={2512.13961},
      archivePrefix={arXiv},
      primaryClass={cs.CL},
      url={https://arxiv.org/abs/2512.13961}, 
}

@misc{yang2025qwen3technicalreport,
      title={Qwen3 Technical Report}, 
      author={An Yang and Anfeng Li and Baosong Yang and Beichen Zhang and Binyuan Hui and Bo Zheng and Bowen Yu and Chang Gao and Chengen Huang and Chenxu Lv and Chujie Zheng and Dayiheng Liu and Fan Zhou and Fei Huang and Feng Hu and Hao Ge and Haoran Wei and Huan Lin and Jialong Tang and Jian Yang and Jianhong Tu and Jianwei Zhang and Jianxin Yang and Jiaxi Yang and Jing Zhou and Jingren Zhou and Junyang Lin and Kai Dang and Keqin Bao and Kexin Yang and Le Yu and Lianghao Deng and Mei Li and Mingfeng Xue and Mingze Li and Pei Zhang and Peng Wang and Qin Zhu and Rui Men and Ruize Gao and Shixuan Liu and Shuang Luo and Tianhao Li and Tianyi Tang and Wenbiao Yin and Xingzhang Ren and Xinyu Wang and Xinyu Zhang and Xuancheng Ren and Yang Fan and Yang Su and Yichang Zhang and Yinger Zhang and Yu Wan and Yuqiong Liu and Zekun Wang and Zeyu Cui and Zhenru Zhang and Zhipeng Zhou and Zihan Qiu},
      year={2025},
      eprint={2505.09388},
      archivePrefix={arXiv},
      primaryClass={cs.CL},
      url={https://arxiv.org/abs/2505.09388}, 
}

@misc{nvidia2025nvidianemotronnano2,
      title={NVIDIA Nemotron Nano 2: An Accurate and Efficient Hybrid Mamba-Transformer Reasoning Model}, 
      author={NVIDIA and : and Aarti Basant and Abhijit Khairnar and Abhijit Paithankar and Abhinav Khattar and Adithya Renduchintala and Aditya Malte and Akhiad Bercovich and Akshay Hazare and Alejandra Rico and Aleksander Ficek and Alex Kondratenko and Alex Shaposhnikov and Alexander Bukharin and Ali Taghibakhshi and Amelia Barton and Ameya Sunil Mahabaleshwarkar and Amy Shen and Andrew Tao and Ann Guan and Anna Shors and Anubhav Mandarwal and Arham Mehta and Arun Venkatesan and Ashton Sharabiani and Ashwath Aithal and Ashwin Poojary and Ayush Dattagupta and Balaram Buddharaju and Banghua Zhu and Barnaby Simkin and Bilal Kartal and Bita Darvish Rouhani and Bobby Chen and Boris Ginsburg and Brandon Norick and Brian Yu and Bryan Catanzaro and Charles Wang and Charlie Truong and Chetan Mungekar and Chintan Patel and Chris Alexiuk and Christian Munley and Christopher Parisien and Dan Su and Daniel Afrimi and Daniel Korzekwa and Daniel Rohrer and Daria Gitman and David Mosallanezhad and Deepak Narayanan and Dima Rekesh and Dina Yared and Dmytro Pykhtar and Dong Ahn and Duncan Riach and Eileen Long and Elliott Ning and Eric Chung and Erick Galinkin and Evelina Bakhturina and Gargi Prasad and Gerald Shen and Haifeng Qian and Haim Elisha and Harsh Sharma and Hayley Ross and Helen Ngo and Herman Sahota and Hexin Wang and Hoo Chang Shin and Hua Huang and Iain Cunningham and Igor Gitman and Ivan Moshkov and Jaehun Jung and Jan Kautz and Jane Polak Scowcroft and Jared Casper and Jian Zhang and Jiaqi Zeng and Jimmy Zhang and Jinze Xue and Jocelyn Huang and Joey Conway and John Kamalu and Jonathan Cohen and Joseph Jennings and Julien Veron Vialard and Junkeun Yi and Jupinder Parmar and Kari Briski and Katherine Cheung and Katherine Luna and Keith Wyss and Keshav Santhanam and Kezhi Kong and Krzysztof Pawelec and Kumar Anik and Kunlun Li and Kushan Ahmadian and Lawrence McAfee and Laya Sleiman and Leon Derczynski and Luis Vega and Maer Rodrigues de Melo and Makesh Narsimhan Sreedhar and Marcin Chochowski and Mark Cai and Markus Kliegl and Marta Stepniewska-Dziubinska and Matvei Novikov and Mehrzad Samadi and Meredith Price and Meriem Boubdir and Michael Boone and Michael Evans and Michal Bien and Michal Zawalski and Miguel Martinez and Mike Chrzanowski and Mohammad Shoeybi and Mostofa Patwary and Namit Dhameja and Nave Assaf and Negar Habibi and Nidhi Bhatia and Nikki Pope and Nima Tajbakhsh and Nirmal Kumar Juluru and Oleg Rybakov and Oleksii Hrinchuk and Oleksii Kuchaiev and Oluwatobi Olabiyi and Pablo Ribalta and Padmavathy Subramanian and Parth Chadha and Pavlo Molchanov and Peter Dykas and Peter Jin and Piotr Bialecki and Piotr Januszewski and Pradeep Thalasta and Prashant Gaikwad and Prasoon Varshney and Pritam Gundecha and Przemek Tredak and Rabeeh Karimi Mahabadi and Rajen Patel and Ran El-Yaniv and Ranjit Rajan and Ria Cheruvu and Rima Shahbazyan and Ritika Borkar and Ritu Gala and Roger Waleffe and Ruoxi Zhang and Russell J. Hewett and Ryan Prenger and Sahil Jain and Samuel Kriman and Sanjeev Satheesh and Saori Kaji and Sarah Yurick and Saurav Muralidharan and Sean Narenthiran and Seonmyeong Bak and Sepehr Sameni and Seungju Han and Shanmugam Ramasamy and Shaona Ghosh and Sharath Turuvekere Sreenivas and Shelby Thomas and Shizhe Diao and Shreya Gopal and Shrimai Prabhumoye and Shubham Toshniwal and Shuoyang Ding and Siddharth Singh and Siddhartha Jain and Somshubra Majumdar and Soumye Singhal and Stefania Alborghetti and Syeda Nahida Akter and Terry Kong and Tim Moon and Tomasz Hliwiak and Tomer Asida and Tony Wang and Tugrul Konuk and Twinkle Vashishth and Tyler Poon and Udi Karpas and Vahid Noroozi and Venkat Srinivasan and Vijay Korthikanti and Vikram Fugro and Vineeth Kalluru and Vitaly Kurin and Vitaly Lavrukhin and Wasi Uddin Ahmad and Wei Du and Wonmin Byeon and Ximing Lu and Xin Dong and Yashaswi Karnati and Yejin Choi and Yian Zhang and Ying Lin and Yonggan Fu and Yoshi Suhara and Zhen Dong and Zhiyu Li and Zhongbo Zhu and Zijia Chen},
      year={2025},
      eprint={2508.14444},
      archivePrefix={arXiv},
      primaryClass={cs.CL},
      url={https://arxiv.org/abs/2508.14444}, 
}

@misc{deepseekai2024deepseekllmscalingopensource,
      title={DeepSeek LLM: Scaling Open-Source Language Models with Longtermism}, 
      author={DeepSeek-AI and : and Xiao Bi and Deli Chen and Guanting Chen and Shanhuang Chen and Damai Dai and Chengqi Deng and Honghui Ding and Kai Dong and Qiushi Du and Zhe Fu and Huazuo Gao and Kaige Gao and Wenjun Gao and Ruiqi Ge and Kang Guan and Daya Guo and Jianzhong Guo and Guangbo Hao and Zhewen Hao and Ying He and Wenjie Hu and Panpan Huang and Erhang Li and Guowei Li and Jiashi Li and Yao Li and Y. K. Li and Wenfeng Liang and Fangyun Lin and A. X. Liu and Bo Liu and Wen Liu and Xiaodong Liu and Xin Liu and Yiyuan Liu and Haoyu Lu and Shanghao Lu and Fuli Luo and Shirong Ma and Xiaotao Nie and Tian Pei and Yishi Piao and Junjie Qiu and Hui Qu and Tongzheng Ren and Zehui Ren and Chong Ruan and Zhangli Sha and Zhihong Shao and Junxiao Song and Xuecheng Su and Jingxiang Sun and Yaofeng Sun and Minghui Tang and Bingxuan Wang and Peiyi Wang and Shiyu Wang and Yaohui Wang and Yongji Wang and Tong Wu and Y. Wu and Xin Xie and Zhenda Xie and Ziwei Xie and Yiliang Xiong and Hanwei Xu and R. X. Xu and Yanhong Xu and Dejian Yang and Yuxiang You and Shuiping Yu and Xingkai Yu and B. Zhang and Haowei Zhang and Lecong Zhang and Liyue Zhang and Mingchuan Zhang and Minghua Zhang and Wentao Zhang and Yichao Zhang and Chenggang Zhao and Yao Zhao and Shangyan Zhou and Shunfeng Zhou and Qihao Zhu and Yuheng Zou},
      year={2024},
      eprint={2401.02954},
      archivePrefix={arXiv},
      primaryClass={cs.CL},
      url={https://arxiv.org/abs/2401.02954}, 
}

@article{deepseek-r1,
   title={DeepSeek-R1 incentivizes reasoning in LLMs through reinforcement learning},
   volume={645},
   ISSN={1476-4687},
   url={http://dx.doi.org/10.1038/s41586-025-09422-z},
   DOI={10.1038/s41586-025-09422-z},
   number={8081},
   journal={Nature},
   publisher={Springer Science and Business Media LLC},
   author={Guo, Daya and Yang, Dejian and Zhang, Haowei and Song, Junxiao and Wang, Peiyi and Zhu, Qihao and Xu, Runxin and Zhang, Ruoyu and Ma, Shirong and Bi, Xiao and Zhang, Xiaokang and Yu, Xingkai and Wu, Yu and Wu, Z. F. and Gou, Zhibin and Shao, Zhihong and Li, Zhuoshu and Gao, Ziyi and Liu, Aixin and Xue, Bing and Wang, Bingxuan and Wu, Bochao and Feng, Bei and Lu, Chengda and Zhao, Chenggang and Deng, Chengqi and Ruan, Chong and Dai, Damai and Chen, Deli and Ji, Dongjie and Li, Erhang and Lin, Fangyun and Dai, Fucong and Luo, Fuli and Hao, Guangbo and Chen, Guanting and Li, Guowei and Zhang, H. and Xu, Hanwei and Ding, Honghui and Gao, Huazuo and Qu, Hui and Li, Hui and Guo, Jianzhong and Li, Jiashi and Chen, Jingchang and Yuan, Jingyang and Tu, Jinhao and Qiu, Junjie and Li, Junlong and Cai, J. L. and Ni, Jiaqi and Liang, Jian and Chen, Jin and Dong, Kai and Hu, Kai and You, Kaichao and Gao, Kaige and Guan, Kang and Huang, Kexin and Yu, Kuai and Wang, Lean and Zhang, Lecong and Zhao, Liang and Wang, Litong and Zhang, Liyue and Xu, Lei and Xia, Leyi and Zhang, Mingchuan and Zhang, Minghua and Tang, Minghui and Zhou, Mingxu and Li, Meng and Wang, Miaojun and Li, Mingming and Tian, Ning and Huang, Panpan and Zhang, Peng and Wang, Qiancheng and Chen, Qinyu and Du, Qiushi and Ge, Ruiqi and Zhang, Ruisong and Pan, Ruizhe and Wang, Runji and Chen, R. J. and Jin, R. L. and Chen, Ruyi and Lu, Shanghao and Zhou, Shangyan and Chen, Shanhuang and Ye, Shengfeng and Wang, Shiyu and Yu, Shuiping and Zhou, Shunfeng and Pan, Shuting and Li, S. S. and Zhou, Shuang and Wu, Shaoqing and Yun, Tao and Pei, Tian and Sun, Tianyu and Wang, T. and Zeng, Wangding and Liu, Wen and Liang, Wenfeng and Gao, Wenjun and Yu, Wenqin and Zhang, Wentao and Xiao, W. L. and An, Wei and Liu, Xiaodong and Wang, Xiaohan and Chen, Xiaokang and Nie, Xiaotao and Cheng, Xin and Liu, Xin and Xie, Xin and Liu, Xingchao and Yang, Xinyu and Li, Xinyuan and Su, Xuecheng and Lin, Xuheng and Li, X. Q. and Jin, Xiangyue and Shen, Xiaojin and Chen, Xiaosha and Sun, Xiaowen and Wang, Xiaoxiang and Song, Xinnan and Zhou, Xinyi and Wang, Xianzu and Shan, Xinxia and Li, Y. K. and Wang, Y. Q. and Wei, Y. X. and Zhang, Yang and Xu, Yanhong and Li, Yao and Zhao, Yao and Sun, Yaofeng and Wang, Yaohui and Yu, Yi and Zhang, Yichao and Shi, Yifan and Xiong, Yiliang and He, Ying and Piao, Yishi and Wang, Yisong and Tan, Yixuan and Ma, Yiyang and Liu, Yiyuan and Guo, Yongqiang and Ou, Yuan and Wang, Yuduan and Gong, Yue and Zou, Yuheng and He, Yujia and Xiong, Yunfan and Luo, Yuxiang and You, Yuxiang and Liu, Yuxuan and Zhou, Yuyang and Zhu, Y. X. and Huang, Yanping and Li, Yaohui and Zheng, Yi and Zhu, Yuchen and Ma, Yunxian and Tang, Ying and Zha, Yukun and Yan, Yuting and Ren, Z. Z. and Ren, Zehui and Sha, Zhangli and Fu, Zhe and Xu, Zhean and Xie, Zhenda and Zhang, Zhengyan and Hao, Zhewen and Ma, Zhicheng and Yan, Zhigang and Wu, Zhiyu and Gu, Zihui and Zhu, Zijia and Liu, Zijun and Li, Zilin and Xie, Ziwei and Song, Ziyang and Pan, Zizheng and Huang, Zhen and Xu, Zhipeng and Zhang, Zhongyu and Zhang, Zhen},
   year={2025},
   month=sept, pages={633–638} }

@misc{lu2022quarkcontrollabletextgeneration,
  title={Quark: Controllable text generation with reinforced unlearning},
  author={Lu, Ximing and Welleck, Sean and Hessel, Jack and Jiang, Liwei and Qin, Lianhui and West, Peter and Ammanabrolu, Prithviraj and Choi, Yejin},
  journal={Advances in neural information processing systems},
  volume={35},
  pages={27591--27609},
  year={2022}
}

@misc{rathi2026shapingcapabilitiestokenleveldata,
      title={Shaping capabilities with token-level data filtering}, 
      author={Neil Rathi and Alec Radford},
      year={2026},
      eprint={2601.21571},
      archivePrefix={arXiv},
      primaryClass={cs.LG},
      url={https://arxiv.org/abs/2601.21571}, 
}

@misc{cobbe2021trainingverifierssolvemath,
      title={Training Verifiers to Solve Math Word Problems}, 
      author={Karl Cobbe and Vineet Kosaraju and Mohammad Bavarian and Mark Chen and Heewoo Jun and Lukasz Kaiser and Matthias Plappert and Jerry Tworek and Jacob Hilton and Reiichiro Nakano and Christopher Hesse and John Schulman},
      year={2021},
      eprint={2110.14168},
      archivePrefix={arXiv},
      primaryClass={cs.LG},
      url={https://arxiv.org/abs/2110.14168}, 
}

@inproceedings{AIME2024,
  author={MAA},
  title = {American Invitational Mathematics Examination - AIME},
  booktitle = {American Invitational Mathematics Examination - AIME 2024},
  year = {2024},
  month = {February},
  url = {https://maa.org/math-competitions/american-invitational-mathematics-examination-aime}
}

@inproceedings{AIME2025,
  author={MAA},
  title = {American Invitational Mathematics Examination - AIME},
  booktitle = {American Invitational Mathematics Examination - AIME 2025},
  year = {2025},
  month = {February},
  url = {https://maa.org/math-competitions/american-invitational-mathematics-examination-aime}
}

@inproceedings{AMC,
  author={MAA},
  title = {American Mathematics Competition - AMC},
  booktitle = {American Mathematics Competition - AMC},
  year = {2023},
  url = {https://maa.org/student-programs/amc/}
}

@misc{hendrycks2021measuringmathematicalproblemsolving,
     title={Measuring Mathematical Problem Solving With the MATH Dataset},
    author={Hendrycks, Dan and Burns, Collin and Kadavath, Saurav and Arora, Akul and Basart, Steven and Tang, Eric and Song, Dawn and Steinhardt, Jacob},
    booktitle={Thirty-fifth Conference on Neural Information Processing Systems Datasets and Benchmarks Track (Round 2)},
    year={2021},
}

@misc{chen2021evaluatinglargelanguagemodels,
      title={Evaluating Large Language Models Trained on Code}, 
      author={Mark Chen and Jerry Tworek and Heewoo Jun and Qiming Yuan and Henrique Ponde de Oliveira Pinto and Jared Kaplan and Harri Edwards and Yuri Burda and Nicholas Joseph and Greg Brockman and Alex Ray and Raul Puri and Gretchen Krueger and Michael Petrov and Heidy Khlaaf and Girish Sastry and Pamela Mishkin and Brooke Chan and Scott Gray and Nick Ryder and Mikhail Pavlov and Alethea Power and Lukasz Kaiser and Mohammad Bavarian and Clemens Winter and Philippe Tillet and Felipe Petroski Such and Dave Cummings and Matthias Plappert and Fotios Chantzis and Elizabeth Barnes and Ariel Herbert-Voss and William Hebgen Guss and Alex Nichol and Alex Paino and Nikolas Tezak and Jie Tang and Igor Babuschkin and Suchir Balaji and Shantanu Jain and William Saunders and Christopher Hesse and Andrew N. Carr and Jan Leike and Josh Achiam and Vedant Misra and Evan Morikawa and Alec Radford and Matthew Knight and Miles Brundage and Mira Murati and Katie Mayer and Peter Welinder and Bob McGrew and Dario Amodei and Sam McCandlish and Ilya Sutskever and Wojciech Zaremba},
      year={2021},
      eprint={2107.03374},
      archivePrefix={arXiv},
      primaryClass={cs.LG},
      url={https://arxiv.org/abs/2107.03374}, 
}

@misc{austin2021programsynthesislargelanguage,
      title={Program Synthesis with Large Language Models}, 
      author={Jacob Austin and Augustus Odena and Maxwell Nye and Maarten Bosma and Henryk Michalewski and David Dohan and Ellen Jiang and Carrie Cai and Michael Terry and Quoc Le and Charles Sutton},
      year={2021},
      eprint={2108.07732},
      archivePrefix={arXiv},
      primaryClass={cs.PL},
      url={https://arxiv.org/abs/2108.07732}, 
}

@misc{hendrycks2021measuringmassivemultitasklanguage,
    title={Measuring Massive Multitask Language Understanding},
    author={Hendrycks, Dan and Burns, Collin and Basart, Steven and Zou, Andy and Mazeika, Mantas and Song, Dawn and Steinhardt, Jacob},
    booktitle={International Conference on Learning Representations},
    year={2021},
}

@misc{wang2024mmluprorobustchallengingmultitask,
      title={MMLU-Pro: A More Robust and Challenging Multi-Task Language Understanding Benchmark},
  author={Wang, Yubo and Ma, Xueguang and Zhang, Ge and Ni, Yuansheng and Chandra, Abhranil and Guo, Shiguang and Ren, Weiming and Arulraj, Aaran and He, Xuan and Jiang, Ziyan and others},
  booktitle={The Thirty-eight Conference on Neural Information Processing Systems Datasets and Benchmarks Track},
      year={2024},
}

@misc{clark2018thinksolvedquestionanswering,
      title={Think you have Solved Question Answering? Try ARC, the AI2 Reasoning Challenge}, 
      author={Peter Clark and Isaac Cowhey and Oren Etzioni and Tushar Khot and Ashish Sabharwal and Carissa Schoenick and Oyvind Tafjord},
      year={2018},
      eprint={1803.05457},
      archivePrefix={arXiv},
      primaryClass={cs.AI},
      url={https://arxiv.org/abs/1803.05457}, 
}

@misc{wettig2024quratingselectinghighqualitydata,
        title={QuRating: selecting high-quality data for training language models},
  author={Wettig, Alexander and Gupta, Aatmik and Malik, Saumya and Chen, Danqi},
  booktitle={Proceedings of the 41st International Conference on Machine Learning},
  pages={52915--52971},
  year={2024}
}

@misc{Kaplan2020Scaling,
  title        = {Scaling Laws for Neural Language Models},
  author       = {Kaplan, Jared and McCandlish, Sam and Henighan, Tom and Brown, Tom B. and Chess, Benjamin and Child, Rewon and Gray, Scott and Radford, Alec and Wu, Jeffrey and Amodei, Dario},
  year         = {2020},
  eprint       = {2001.08361},
  archivePrefix= {arXiv},
  primaryClass = {cs.LG},
  doi          = {10.48550/arXiv.2001.08361}
}

@article{ankner2024critique,
  title={Critique-out-loud reward models},
  author={Ankner, Zachary and Paul, Mansheej and Cui, Brandon and Chang, Jonathan D and Ammanabrolu, Prithviraj},
  journal={arXiv preprint arXiv:2408.11791},
  year={2024}
}

@inproceedings{
zhang2025generative,
title={Generative Verifiers: Reward Modeling as Next-Token Prediction},
author={Lunjun Zhang and Arian Hosseini and Hritik Bansal and Mehran Kazemi and Aviral Kumar and Rishabh Agarwal},
booktitle={The Thirteenth International Conference on Learning Representations},
year={2025},
url={https://openreview.net/forum?id=Ccwp4tFEtE}
}

@misc{tan2026selfimprovingpretrainingusingposttrained,
      title={Self-Improving Pretraining: using post-trained models to pretrain better models}, 
      author={Ellen Xiaoqing Tan and Jack Lanchantin and Shehzaad Dhuliawala and Danwei Li, Thao Nguyen and Jing Xu and Ping Yu and Ilia Kulikov and Sainbayar Sukhbaatar and Jason Weston and Xian Li and Olga Golovneva},
      year={2026},
      eprint={2601.21343},
      archivePrefix={arXiv},
      primaryClass={cs.CL},
      url={https://arxiv.org/abs/2601.21343}, 
}

@inproceedings{
blakeney2024does,
title={Does your data spark joy? Performance gains from domain upsampling at the end of training},
author={Cody Blakeney and Mansheej Paul and Brett W. Larsen and Sean Owen and Jonathan Frankle},
booktitle={Workshop on Efficient Systems for Foundation Models II @ ICML2024},
year={2024},
url={https://openreview.net/forum?id=g5CK5ReB3k}
}

@article{bradley-terry,
 ISSN = {00063444, 14643510},
 URL = {http://www.jstor.org/stable/2334029},
 author = {Ralph Allan Bradley and Milton E. Terry},
 journal = {Biometrika},
 number = {3/4},
 pages = {324--345},
 publisher = {[Oxford University Press, Biometrika Trust]},
 title = {Rank Analysis of Incomplete Block Designs: I. The Method of Paired Comparisons},
 urldate = {2026-04-28},
 volume = {39},
 year = {1952}
}

@article{Liu1989LBFGS,
  author    = {Dong C. Liu and Jorge Nocedal},
  title     = {On the limited memory BFGS method for large scale optimization},
  journal   = {Mathematical Programming},
  volume    = {45},
  pages     = {503--528},
  year      = {1989},
  doi       = {10.1007/BF01589116}
}

@misc{he2024olympiadbenchchallengingbenchmarkpromoting,
      title={OlympiadBench: A Challenging Benchmark for Promoting AGI with Olympiad-Level Bilingual Multimodal Scientific Problems}, 
      author={Chaoqun He and Renjie Luo and Yuzhuo Bai and Shengding Hu and Zhen Leng Thai and Junhao Shen and Jinyi Hu and Xu Han and Yujie Huang and Yuxiang Zhang and Jie Liu and Lei Qi and Zhiyuan Liu and Maosong Sun},
      year={2024},
      eprint={2402.14008},
      archivePrefix={arXiv},
      primaryClass={cs.CL},
      url={https://arxiv.org/abs/2402.14008}, 
}

@misc{jain2024livecodebenchholisticcontaminationfree,
      title={LiveCodeBench: Holistic and Contamination Free Evaluation of Large Language Models for Code}, 
      author={Naman Jain and King Han and Alex Gu and Wen-Ding Li and Fanjia Yan and Tianjun Zhang and Sida Wang and Armando Solar-Lezama and Koushik Sen and Ion Stoica},
      year={2024},
      eprint={2403.07974},
      archivePrefix={arXiv},
      primaryClass={cs.SE},
      url={https://arxiv.org/abs/2403.07974}, 
}

@misc{lewkowycz2022solvingquantitativereasoningproblems,
      title={Solving Quantitative Reasoning Problems with Language Models}, 
      author={Aitor Lewkowycz and Anders Andreassen and David Dohan and Ethan Dyer and Henryk Michalewski and Vinay Ramasesh and Ambrose Slone and Cem Anil and Imanol Schlag and Theo Gutman-Solo and Yuhuai Wu and Behnam Neyshabur and Guy Gur-Ari and Vedant Misra},
      year={2022},
      eprint={2206.14858},
      archivePrefix={arXiv},
      primaryClass={cs.CL},
      url={https://arxiv.org/abs/2206.14858}, 
}

@article{zhou2023instruction,
  title         = {Instruction-Following Evaluation for Large Language Models},
  author        = {Jeffrey Zhou and Tianjian Lu and Swaroop Mishra and Siddhartha Brahma and Sujoy Basu and Yi Luan and Denny Zhou and Le Hou},
  journal       = {arXiv preprint arXiv:2311.07911},
  year          = {2023},
  doi           = {10.48550/arXiv.2311.07911},
  eprint        = {2311.07911},
  archivePrefix = {arXiv}
}

@misc{hu2024minicpmunveilingpotentialsmall,
      title={MiniCPM: Unveiling the Potential of Small Language Models with Scalable Training Strategies}, 
      author={Shengding Hu and Yuge Tu and Xu Han and Chaoqun He and Ganqu Cui and Xiang Long and Zhi Zheng and Yewei Fang and Yuxiang Huang and Weilin Zhao and Xinrong Zhang and Zheng Leng Thai and Kaihuo Zhang and Chongyi Wang and Yuan Yao and Chenyang Zhao and Jie Zhou and Jie Cai and Zhongwu Zhai and Ning Ding and Chao Jia and Guoyang Zeng and Dahai Li and Zhiyuan Liu and Maosong Sun},
      year={2024},
      eprint={2404.06395},
      archivePrefix={arXiv},
      primaryClass={cs.CL},
      url={https://arxiv.org/abs/2404.06395}, 
}

@software{Leo_Gao_Jonathan, title={A framework for few-shot language model evaluation}, DOI={10.5281/zenodo.5371629}, publisher={Zenodo}, author={Leo Gao and Jonathan Tow and Biderman, Stella and Black, Sid and DiPofi, Anthony and Foster, Charles and Golding, Laurence and Hsu, Jeffrey and McDonell, Kyle and Muennighoff, Niklas}, year={2021} }

@misc{liu2023codegeneratedchatgptreally,
      title={Is Your Code Generated by ChatGPT Really Correct? Rigorous Evaluation of Large Language Models for Code Generation}, 
      author={Jiawei Liu and Chunqiu Steven Xia and Yuyao Wang and Lingming Zhang},
      year={2023},
      eprint={2305.01210},
      archivePrefix={arXiv},
      primaryClass={cs.SE},
      url={https://arxiv.org/abs/2305.01210}, 
}

@article{LaiRACE2017,
  author       = {Guokun Lai and
                  Qizhe Xie and
                  Hanxiao Liu and
                  Yiming Yang and
                  Eduard H. Hovy},
  title        = {{RACE:} Large-scale ReAding Comprehension Dataset From Examinations},
  journal      = {CoRR},
  volume       = {abs/1704.04683},
  year         = {2017},
  url          = {http://arxiv.org/abs/1704.04683},
  eprinttype   = {arXiv},
  eprint       = {1704.04683},
  bibsource    = {dblp computer science bibliography, https://dblp.org}
}

@misc{ainslie2023gqatraininggeneralizedmultiquery,
      title={GQA: Training Generalized Multi-Query Transformer Models from Multi-Head Checkpoints}, 
      author={Joshua Ainslie and James Lee-Thorp and Michiel de Jong and Yury Zemlyanskiy and Federico Lebrón and Sumit Sanghai},
      year={2023},
      eprint={2305.13245},
      archivePrefix={arXiv},
      primaryClass={cs.CL},
      url={https://arxiv.org/abs/2305.13245}, 
}

@misc{so2022primersearchingefficienttransformers,
      title={Primer: Searching for Efficient Transformers for Language Modeling}, 
      author={David R. So and Wojciech Mańke and Hanxiao Liu and Zihang Dai and Noam Shazeer and Quoc V. Le},
      year={2022},
      eprint={2109.08668},
      archivePrefix={arXiv},
      primaryClass={cs.LG},
      url={https://arxiv.org/abs/2109.08668}, 
}

@misc{loshchilov2019decoupledweightdecayregularization,
      title={Decoupled Weight Decay Regularization}, 
      author={Ilya Loshchilov and Frank Hutter},
      year={2019},
      eprint={1711.05101},
      archivePrefix={arXiv},
      primaryClass={cs.LG},
      url={https://arxiv.org/abs/1711.05101}, 
}

@misc{guha2025openthoughtsdatarecipesreasoning,
      title={OpenThoughts: Data Recipes for Reasoning Models}, 
      author={Etash Guha and Ryan Marten and Sedrick Keh and Negin Raoof and Georgios Smyrnis and Hritik Bansal and Marianna Nezhurina and Jean Mercat and Trung Vu and Zayne Sprague and Ashima Suvarna and Benjamin Feuer and Liangyu Chen and Zaid Khan and Eric Frankel and Sachin Grover and Caroline Choi and Niklas Muennighoff and Shiye Su and Wanjia Zhao and John Yang and Shreyas Pimpalgaonkar and Kartik Sharma and Charlie Cheng-Jie Ji and Yichuan Deng and Sarah Pratt and Vivek Ramanujan and Jon Saad-Falcon and Jeffrey Li and Achal Dave and Alon Albalak and Kushal Arora and Blake Wulfe and Chinmay Hegde and Greg Durrett and Sewoong Oh and Mohit Bansal and Saadia Gabriel and Aditya Grover and Kai-Wei Chang and Vaishaal Shankar and Aaron Gokaslan and Mike A. Merrill and Tatsunori Hashimoto and Yejin Choi and Jenia Jitsev and Reinhard Heckel and Maheswaran Sathiamoorthy and Alexandros G. Dimakis and Ludwig Schmidt},
      year={2025},
      eprint={2506.04178},
      archivePrefix={arXiv},
      primaryClass={cs.LG},
      url={https://arxiv.org/abs/2506.04178}, 
}
